# CNN-Based Retinal OCT Classification: Research Gap Analysis and a From-Scratch Baseline, ED126-A

Shkëlqim Sherifi
*Dept. of Computer Science*
*University of Tetovo*
Tetove, R.N.Macedonia
0009-0006-2227-5533

Festim Halili
*Dept. of Computer Science*
*University of Tetovo*
Tetove, R.N.Macedonia
0000-0001-7043-2173

Merita Kasa-Halili
*Dept. of Computer Science*
*University of Tetovo*
Tetove, R.N.Macedonia
0009-0001-9778-8820

***Abstract*- Automated classification of retinal pathologies from Optical Coherence Tomography (OCT) images remains challenging because diagnostically relevant morphological patterns may exhibit substantial inter-class similarity. Although high classification accuracies have been reported on benchmark OCT datasets, cross-study comparison is limited by heterogeneous evaluation protocols and by the limited characterization of compact convolutional neural network (CNN) baselines trained from random initialization. This study addresses this gap through two contributions. First, it provides a focused review of CNN-based retinal OCT classification and identifies methodological considerations relevant to controlled benchmarking. Second, it presents ED126-A, a purpose-built four-class CNN baseline trained from scratch under a documented experimental protocol on the Kermany labeled OCT dataset. The model contains 1,241,380 total parameters, including 1,239,460 trainable parameters, and is evaluated using training-dynamics analysis, class-wise performance metrics, confusion-matrix analysis, confidence analysis, t-SNE representation-space visualization, Grad-CAM, hierarchical feature-map analysis, and one-vs-rest ROC/Precision-Recall analysis. On the image-level held-out test set, ED126-A achieved 95.03% accuracy, with a macro-F1 of 0.9205 and a weighted-F1 of 0.9521. The principal class-conditional limitation occurred for Drusen, which attained a recall of 0.9605 but a precision of 0.7493 and an F1-score of 0.8419. The confusion matrix identified Normal to Drusen and CNV to Drusen as major error channels. These results support the use of ED126-A as a documented from-scratch reference baseline and provide an empirical basis for controlled evaluation of transfer learning, increased depth, attention, multi-scale feature fusion, and calibration in subsequent studies.**

## I. INTRODUCTION

Optical Coherence Tomography (OCT) is a high-resolution, non-invasive imaging modality that provides cross-sectional visualization of retinal structure and enables detailed characterization of morphological changes associated with retinal disease [1], [2], [3], [4], [5]. In particular, OCT is widely used to visualize structural manifestations associated with conditions such as Choroidal Neovascularization (CNV), Diabetic Macular Edema (DME), and Drusen [4], [5], [6]. The availability of large labeled OCT datasets has consequently established retinal OCT classification as an important benchmark problem for automated medical image analysis.

Despite the structural information available in OCT images, reliable interpretation remains challenging because diagnostically relevant patterns may be subtle, spatially localized, and partially overlapping across disease categories [1], [7]. Variability in image acquisition, artifacts, and speckle-related appearance further complicate interpretation, while visually similar manifestations may occur across distinct pathological classes. These characteristics make automated OCT classification a demanding problem in which high aggregate accuracy alone may not adequately characterize class-specific diagnostic behavior.

Convolutional Neural Networks (CNNs) have become a standard approach for retinal image analysis because they learn hierarchical feature representations directly from image data, replacing manually designed feature descriptors with learned convolutional representations [8], [9], [10], [11]. The broader evolution of deep-learning architectures for visual representation and image classification, from multilayer perceptron's to convolutional neural networks and hybrid models, has also been examined in recent architectural surveys [12]. In retinal OCT classification, both a relatively compact, purpose-built CNN (1.24 million parameters) and larger transfer-learning architectures have demonstrated strong performance on established benchmarks. However, reported results are often associated with different architectural designs, training protocols, class distributions, and evaluation procedures, limiting direct attribution of performance differences to individual architectural choices.

Although numerous CNN-based OCT classifiers report high benchmark accuracy, comparatively limited attention has been given to establishing compact, purpose-built baselines under fully documented and controlled experimental protocols. Without such a reference point, the incremental contribution of subsequent architectural enrichment, including ImageNet pretraining, attention mechanisms, or multi-scale feature fusion, cannot be cleanly separated from differences in data handling, optimization, parameterization, or classification-head design. Moreover, aggregate accuracy and weighted metrics may obscure class-conditional failure modes, particularly when the class distribution is heterogeneous. A baseline that reports detailed per-class performance and error structure is therefore useful for establishing a more interpretable reference for subsequent model comparisons.

This study addresses the identified gap by designing and evaluating ED126-A, a purpose-built CNN for four-class retinal OCT classification involving CNV, DME, Drusen, and Normal categories. The model is trained end-to-end from random initialization under a fully documented experimental protocol, without transfer learning or model ensembling. The architecture contains 1,241,380 total parameters, of which 1,239,460 are trainable, and is evaluated on an image-level held-out test set using complementary quantitative and representation-based analyses. The objective is not to claim state-of-the-art performance, but to provide a documented reference baseline whose class-conditional behavior can serve as a controlled point of comparison for subsequent architectural studies.

The study makes two principal contributions. First, it provides a focused review of CNN-based retinal OCT classification and identifies methodological considerations relevant to controlled baseline construction and comparative evaluation. Second, it presents ED126-A as a compact, from-scratch baseline and characterizes its behavior through multi-faceted evaluation, with particular emphasis on class-conditional performance, recurrent confusion patterns, representation-space structure, confidence behavior, and interpretability. The resulting baseline is intended to support controlled evaluation of transfer learning and architectural enrichment in subsequent studies.

The research questions addressed are:

**RQ1**: Does a purpose-built CNN trained from scratch on the Kermany OCT dataset provide a useful empirical baseline for four-class OCT classification?

**RQ2**: What are the class-conditional performance characteristics of ED126-A, and which pathological classes constitute the principal sources of residual classification error?

The remainder of this paper is organized as follows: **Section II** provides a focused review of CNN-based retinal OCT classification and identifies methodological considerations relevant to controlled baseline construction. **Section III** describes the experimental methodology, including the dataset, data partitioning, preprocessing and augmentation, ED126-A architecture, training configuration, and evaluation protocol. **Section IV** presents the experimental results. **Section V** discusses the principal findings, their interpretation, implications for subsequent benchmarking, and study limitations. **Section VI** concludes the paper.

## II. Related Work

Convolutional neural networks have become a widely used framework for automated retinal OCT classification because they can learn hierarchical image representations directly from training data. Studies on the Kermany OCT benchmark have demonstrated that both conventional CNNs and more elaborate architectures can achieve high classification performance across CNV, DME, Drusen, and Normal categories [3], [5], [6], [13], [14]. These results establish the feasibility of automated four-class OCT classification while also illustrating substantial variation in architectural depth, parameterization, optimization strategy, and evaluation protocol across studies. A recent scoping review of 55 research records (2017-2026) identified five key gaps in CNN-based retinal OCT classification: controlled from-scratch baselines, ImageNet-to-OCT transfer evaluation, attention and multi-scale representations, learned representation analysis, and cross-scanner generalization [15].

Compact architectures have received particular attention in retinal OCT analysis because they offer the potential for reduced computational and memory requirements while maintaining high benchmark performance. OctNET, a lightweight six-block CNN, reported 99.69% accuracy on the four-class Kermany task while substantially reducing parameter count relative to ResNet-50 [3]. A lightweight CNN incorporating adversarial robustness evaluation reported approximately 94-96% accuracy under 10-fold cross-validation and incorporated LIME and SHAP for interpretability [5]. Another compact six-layer CNN reported 94.87% accuracy within a 2 MB model footprint and incorporated LIME and Grad-CAM [16]. OCT-Caps, based on a comparatively shallow capsule-network formulation, reported a mean accuracy of 99.6% while emphasizing preservation of spatial context [17]. Collectively, these studies demonstrate that substantial performance can be obtained with architectures considerably smaller than conventional ImageNet-scale networks.

In parallel with compact CNN development, retinal OCT classification has increasingly incorporated transfer learning and architectural enrichment, including pretrained convolutional backbones, attention mechanisms, and hybrid or multi-scale feature representations [6], [9], [11], [14]. Such approaches can increase representational capacity or introduce inductive biases intended to emphasize diagnostically relevant structures. However, comparisons across these studies are complicated when differences in initialization, architecture, preprocessing, optimization, class weighting, or evaluation protocol occur simultaneously. Consequently, high performance reported by an enriched architecture does not, by itself, quantify the incremental contribution of the enrichment relative to a matched from-scratch baseline.

Despite the strong aggregate performance reported by compact and enriched OCT classifiers, an important methodological gap remains. Published studies frequently emphasize overall accuracy or aggregate metrics, providing limited information about class-conditional failure modes and the specific sources of residual error. In addition, differences in architecture, initialization, data partitioning, class weighting, optimization, and classification-head design can prevent direct attribution of performance gains to individual architectural components. A compact, purpose-built CNN trained from random initialization under a fully documented and fixed protocol can therefore serve as a useful reference condition for subsequent comparative experiments.

The present study addresses this gap through ED126-A, a purpose-built four-class CNN trained from scratch and evaluated using a standardized experimental protocol. Rather than competing directly for state-of-the-art accuracy, ED126-A is intended to provide a documented reference baseline and to characterize its class-conditional behavior through complementary quantitative and representation-based analyses. This baseline provides the experimental foundation for subsequent controlled evaluation of transfer learning, increased depth, attention, multi-scale feature fusion, and calibration.

## III. Methodology

### A. Research Design and Experimental Objective

This study adopts a quantitative empirical design centered on the construction and evaluation of ED126-A, a purpose-built convolutional neural network trained from random initialization for four-class retinal OCT classification. The primary objective is to establish a documented reference baseline under a fixed experimental protocol and to characterize its class-conditional performance and residual error structure. The experimental design deliberately excludes transfer learning, architectural enrichment, and model ensembling so that the baseline can serve as a reference condition for subsequent controlled comparisons.

## B. Dataset and Data Partitioning

Table I summarizes four retinal image datasets considered during dataset selection.

TABLE I. COMPARISON OF RETINAL IMAGE DATASETS

| ***Attribute*** | ***EyeNet[18]*** | ***STARE[19]*** | ***DRIVE[20]*** | ***OCT[21]*** |
|---|---|---|---|---|
| ***Dataset name*** | EyeNet (clinical label collection for ophthalmology) | STARE (Structured Analysis of the Retina) | DRIVE (Digital Retinal Images for Vessel Extraction) | Optical Coherence Tomography - OCT (Retinal) |
| ***Modality*** | Color fundus (retinal) photographs | Color fundus photographs | Color fundus photographs | Retinal OCT B-scans (cross-sectional) |
| ***Images/splits*** | 70% train; 10% val; 20% test | 20 images (10 normal + 10 diseased). No formal train/val/test split | 40 images total: 20 training + 20 test | 109,309 OCT grayscale images. 70% train; 20% val; 10% test. |
| ***Resolution/format*** | Original resolution unspecified. Image format reported as JPEG/PNG. | 700×605 px. PPM format. | 565×584 px. TIFF/PNG formats. | ~512×496 px OCT B-Scan. JPEG 8-bit format. |
| ***Labels/annotations*** | 32 retinal disease labels per image. No segmentation or bounding-box labels. | Vessel masks (binary) from two annotators. No segmentation or bounding-box labels. | Vessel masks (binary) from an expert. Optic disc masks. No segmentation or bounding-box labels. | 4 classes (CNV, DME, Drusen, Normal). No segmentation or bounding-box labels. |
| ***Patient/sample count, demographics*** | Unspecified. Images from multiple patients via RIB | 20 distinct eyes (patients). Diagnoses include various retinopathies; half have pathology. No age/sex info given. | 40 eyes (20 train, 20 test) from diabetic screening; patients are older adults. | 2,442 patients from Zhongshan Ophthalmic Center (Guangzhou), patients were older Asian adults. |
| ***Typical tasks/benchmarks*** | Multi-class classification | Vessel segmentation | Vessel segmentation | Classification of macular disease |
| ***Strengths*** | Diversity: 32 clinical conditions, large variety | Precise segmentation: Dual expert annotations for vessels | Standard benchmark: Well-known, pre-split dataset | Large scale: more than 109k images, well-labeled for 4 classes. |
| **Limitations** | Unbalanced classes: Some classes (rare diseases) have very few images. | Exceedingly small: Only 20 labeled images, so prone to overfitting if used alone. | Small: Just 40 images total. | OCT-only: Focuses on macular diseases, no fundus images. |

- ***B.1. Dataset selection and benchmark context.*** Several publicly available retinal image datasets support different ophthalmic analysis tasks, including fundus-image classification and vessel segmentation. Table I summarizes the principal characteristics of four datasets considered during dataset selection. Because the objective of the present study is four-class retinal OCT classification, the Labeled Optical Coherence Tomography (OCT) dataset introduced by Kermany et al. (2018) [22] was selected as the experimental benchmark.

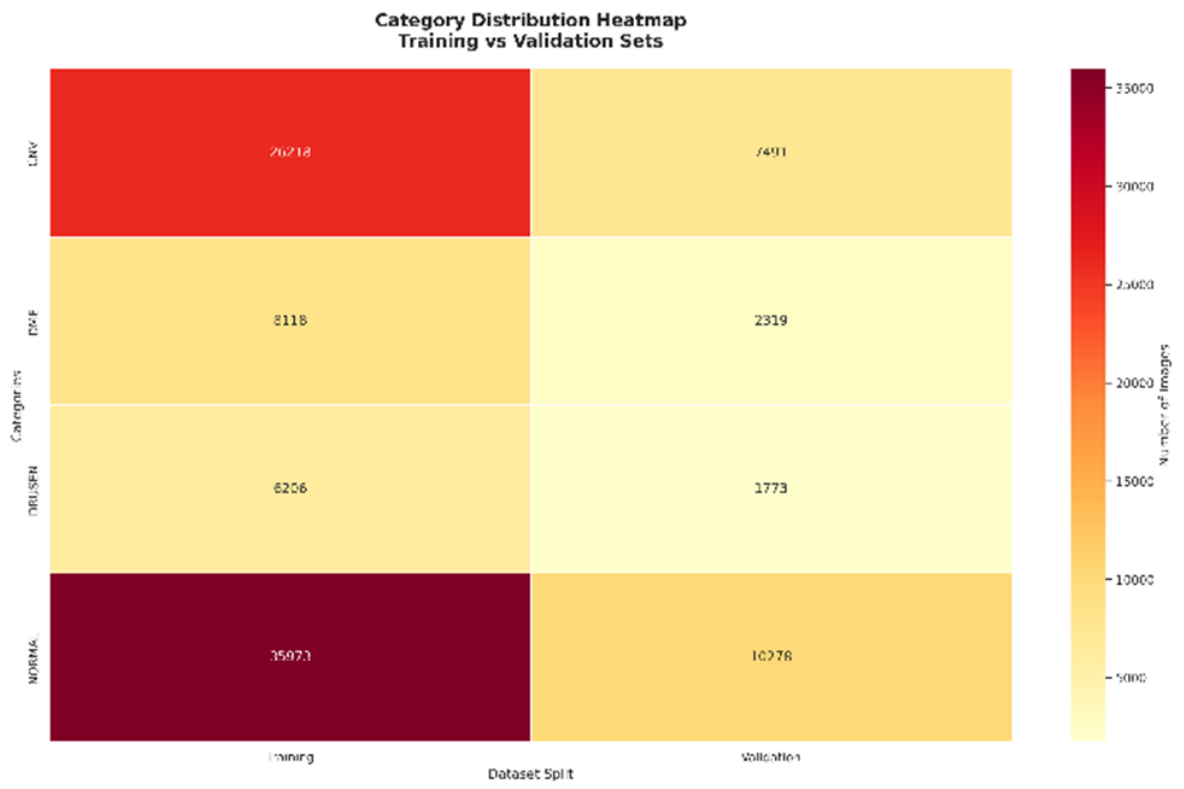

Fig. 1. Distribution of training and validation images across the four retinal OCT classes.

- ***B.2. Dataset composition and class definitions.*** The selected dataset contains 109,309 grayscale OCT images distributed across four diagnostic categories: CNV, DME, Drusen, and Normal [22]. The dataset is widely used as a benchmark for retinal OCT classification and provides a large-scale four-class setting for evaluating CNN-based approaches.

*Class Description*. The four diagnostic categories are:

*Choroidal Neovascularization (CNV)*: Pathological proliferation of new blood vessels originating from the choroid, penetrating Bruch's membrane into the sub-retinal pigment epithelium space.

*Diabetic Macular Edema (DME)*: Fluid accumulation within the macular region secondary to vascular leakage from diabetic retinopathy, disrupting the photoreceptor layer.

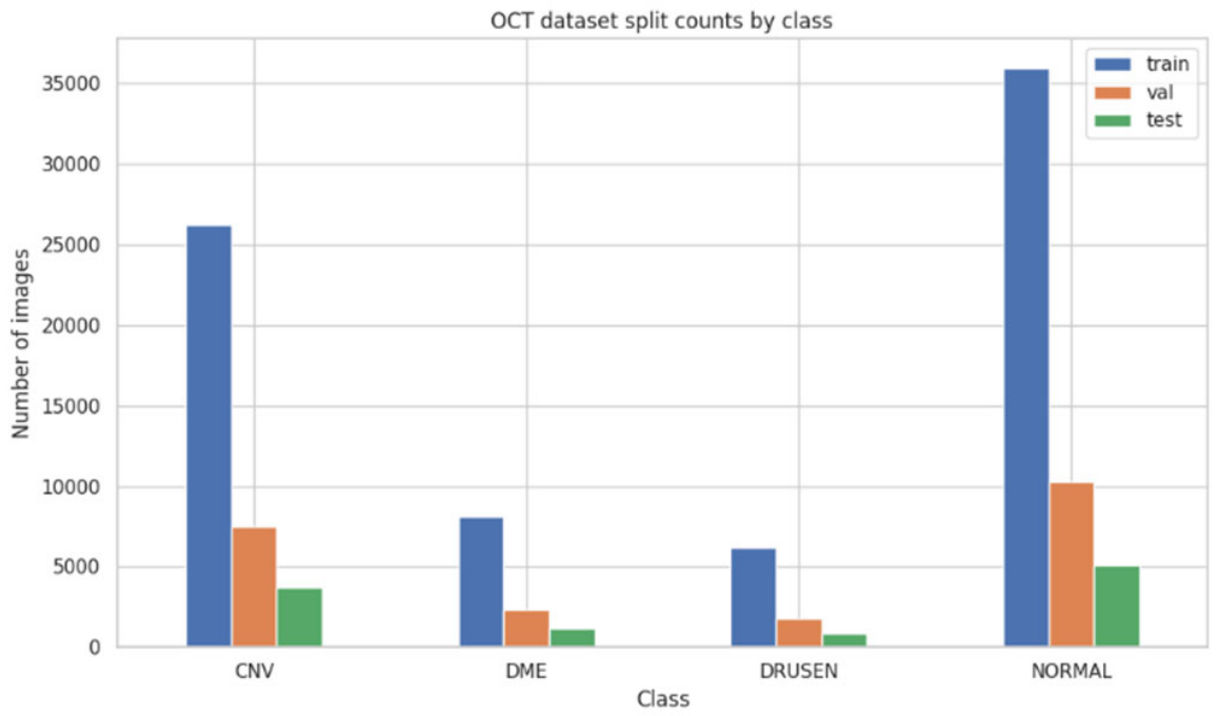

Fig. 2. Class distribution of the retinal OCT dataset across CNV, DME, Drusen, and Normal categories.

*Drusen*: Small, hyaline deposits with well-delineated margins; individually low-risk but recognized as prodromal indicators of progressive retinal disease.

*Normal*: Healthy retinal tissue with intact laminar architecture.

- ***B.3. Dataset partitioning.*** The experimental partition followed the organization provided by [21],

comprising 76,515 training images, 21,861 validation images, and 10,933 test images, corresponding approximately to a 70%/20%/10% split. The train/validation/test partition was performed at the image level rather than at the patient level. Consequently, images from the same patient may occur in more than one partition. The reported performance should therefore be interpreted as performance under the specified image-level partitioning protocol and not as evidence of patient-independent generalization.

TABLE II. DATASET STATISTICS

| Dataset Train/Validation | | |
|---|---|---|
| ***Statistic*** | ***Training Set*** | ***Validation Set*** |
| Unique categories | 4 | 4 |
| Total Images | 76515 | 21861 |
| Mean | 19128 | 5465 |
| Median | 17168 | 4905 |
| Standard Deviation | 12472 | 3563 |
| Min | 6206 | 1773 |
| Max | 35973 | 10278 |
| Coefficient of variation | 0.652 | 0.652 |

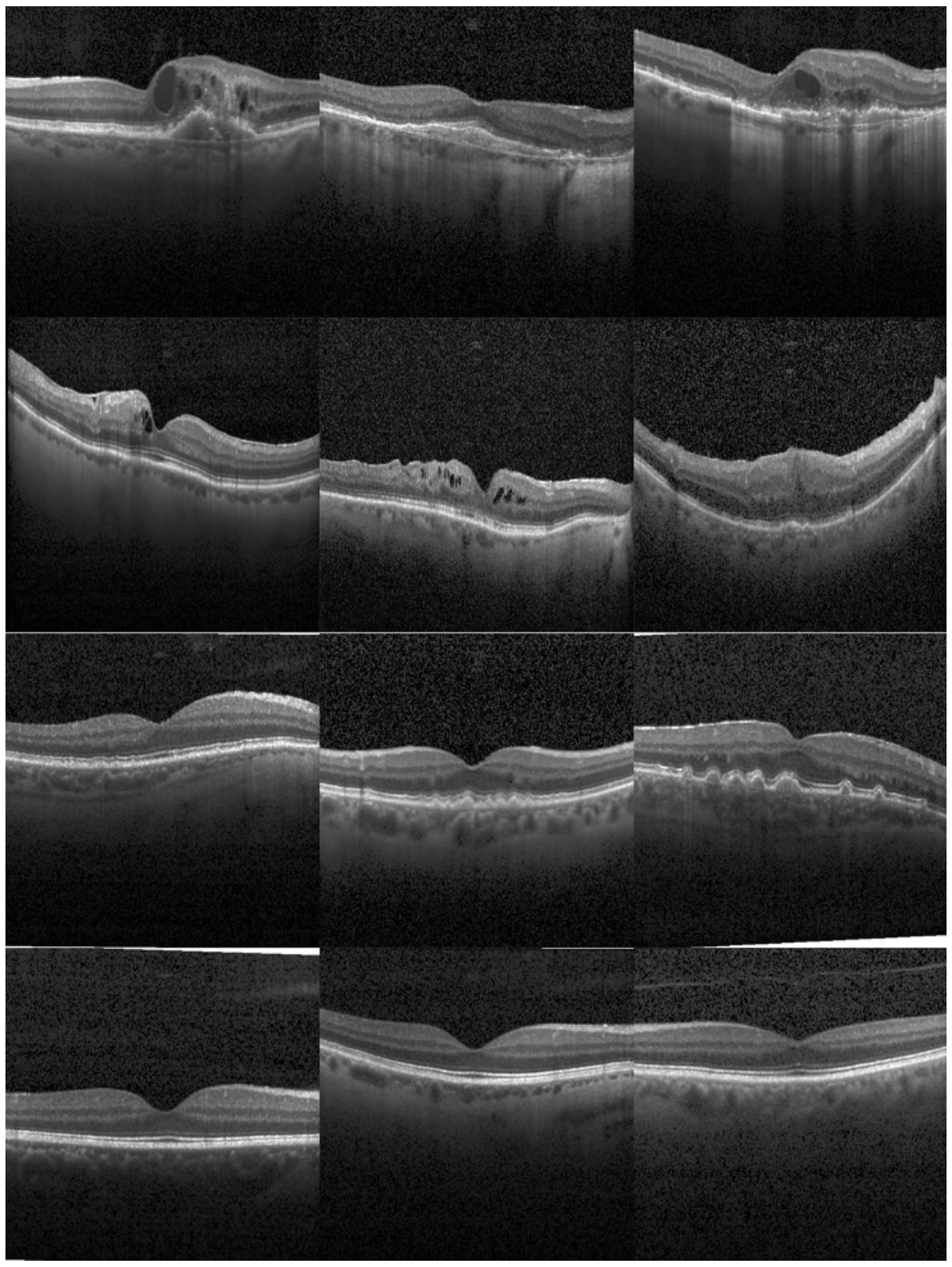

Fig. 3. Distribution of test-set images across the four retinal OCT classes.

- ***B.4. Class distribution.*** The four classes exhibit substantial count heterogeneity. The coefficient of variation across the class counts is approximately 65%, with the Normal category substantially larger than the disease categories. This imbalance was incorporated into the training strategy through balanced class weighting, described in Section III-E.

### C. Preprocessing and Data Augmentation

- ***C.1. Image preprocessing.*** All input images were standardized to 224 × 224 pixels and represented as single-channel grayscale tensors. Pixel intensities were converted from the original 8-bit range [0, 255] to normalized values in [0, 1] by division by 255. The resulting input tensor has shape (224, 224, 1).

**1) DATA AUGMENTATION**

```
Input
RECEIVE input_image (grayscale, uint8, 224×224×1)
Random Horizontal Flip
SAMPLE u ∈ [0, 1]
IF u < 0.5 THEN
     APPLY HORIZONTAL FLIP to input_image.
ENDIF
Random Rotation
SAMPLE angle ∈ [-0.05 x 360°, +0.05 x 360°]
APPLY ROTATION by angle to input_image.
     fill_mode = "nearest"
Random Translation
SAMPLE shift_h ∈ [-0.05 x H, +0.05 x H]
SAMPLE shift_w ∈ [-0.05 x W, +0.05 x W]
APPLY TRANSLATION (shift_h, shift_w) to input_image.
     fill_mode = "nearest"
Random Zoom
SAMPLE zoom_h ∈ [-0.1, +0.1]
SAMPLE zoom_w ∈ [-0.1, +0.1]
COMPUTE new_H = ROUND (H × (1 + zoom_h))
COMPUTE new_W = ROUND (W × (1 + zoom_w))
RESIZE input_image to (new_H, new_W) then CENTER-CROP to (H, W)
     fill_mode = "nearest"
Random Contrast
SAMPLE contrast_factor ∈ [0.85, 1.15]
COMPUTE mean_pixel = MEAN (input_image)
APPLY image ← (image − mean_pixel) × contrast_factor + mean_pixel.
Normalization (Rescaling)
CONVERT image to TENSOR (float32)
APPLY pixel ÷ 255.0 → values ∈ [0.0, 1.0]
-layer name = "rescale_0_1"
Output
RETURN normalized tensor (float32, 224×224×1, values ∈ [0.0, 1.0])
```

Code 1 Pseudo-code Data Augmentation

- ***C.2. Augmentation.*** During training, stochastic augmentation was applied using horizontal flipping, small-angle rotation, translation, zoom, and contrast adjustment. The transformations were implemented as non-trainable preprocessing layers within the model graph and were active only during training. Validation and test images were evaluated without stochastic augmentation.

Fig. 5 introduces representative samples from a single training batch of original grayscale images at 224 × 224 resolution across four disease classes, and Fig. 6 shows the same images after stochastic augmentation: subtle horizontal flips, rotations, translations, zoom, and contrast shifts while the morphology is preserved.

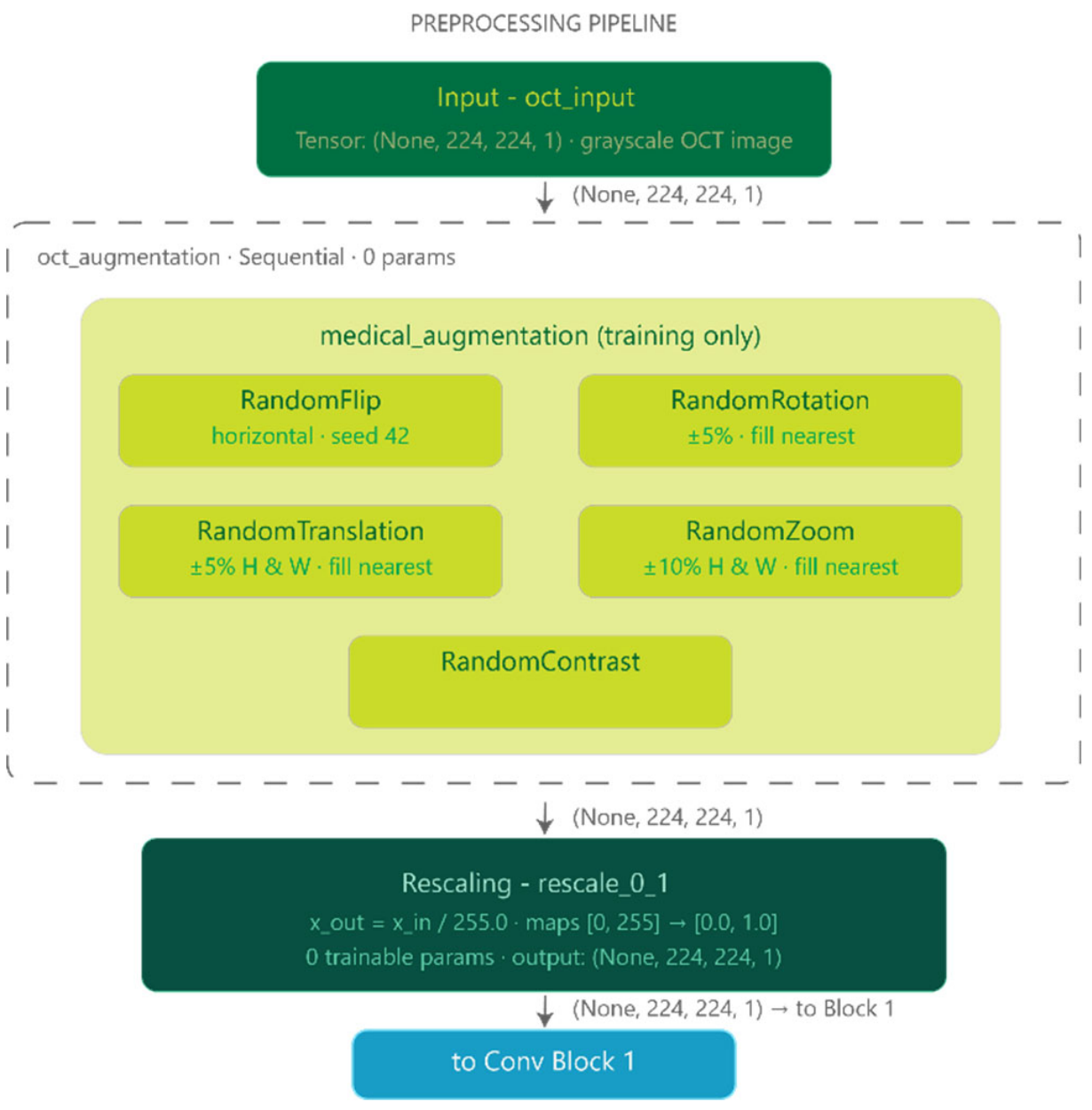

Fig. 4. Preprocessing and training-time augmentation pipeline for retinal OCT images.

- *Random Horizontal Flip* (p=0.5, seed=42). Horizontal flipping was used as a data augmentation to introduce lateral reflection invariance while preserving the principal cross-sectional retinal morphology represented in the images. Vertical flipping was deliberately excluded because anatomical orientation in clinical eye images is predominantly fixed, making vertical inversion an unrealistic transformation that could introduce misleading geometric distortions.

- *Random Rotation* ( $\pm 5\%$ , approximately $\pm 18°$ , $fill = nearest$ ). Small-angle rotations simulate acquisition tilt without introducing out-of-distribution backgrounds, as nearest-neighbor fill avoids black border artifacts.

- *Random Translation* ( $\pm 5\%\ of\ H\ and\ W, fill = nearest$ ). Modest translational perturbation addresses variability in OCT scan centering across acquisition devices and operations.

- *Random Zoom* ($\pm$10% of H and W, fill = nearest). Zoom augmentation introduces moderate scale variation that may approximate differences in scan magnification, field of view, or retinal sampling across acquisitions, in both height and width directions - the image is resized to the new dimensions and then center-cropped back to the original 224 $x$ 224 resolution, emulating variation in imaging distance or magnification.

- *Random Contrast (factor 0.15, seed 42).* Random contrast adjustment computes the per-image mean pixel intensity and scales the deviation from that mean by a sampled contrast factor in the range [0.85, 1.15], thereby introducing mild contrast variation while preserving the image mean.

- *Rescaling.* Pixel intensities are linearly rescaled from the 8-bit range [0,255] to [0,1] using ($x_{\text{out}} = x_{\text{in}}/255$).

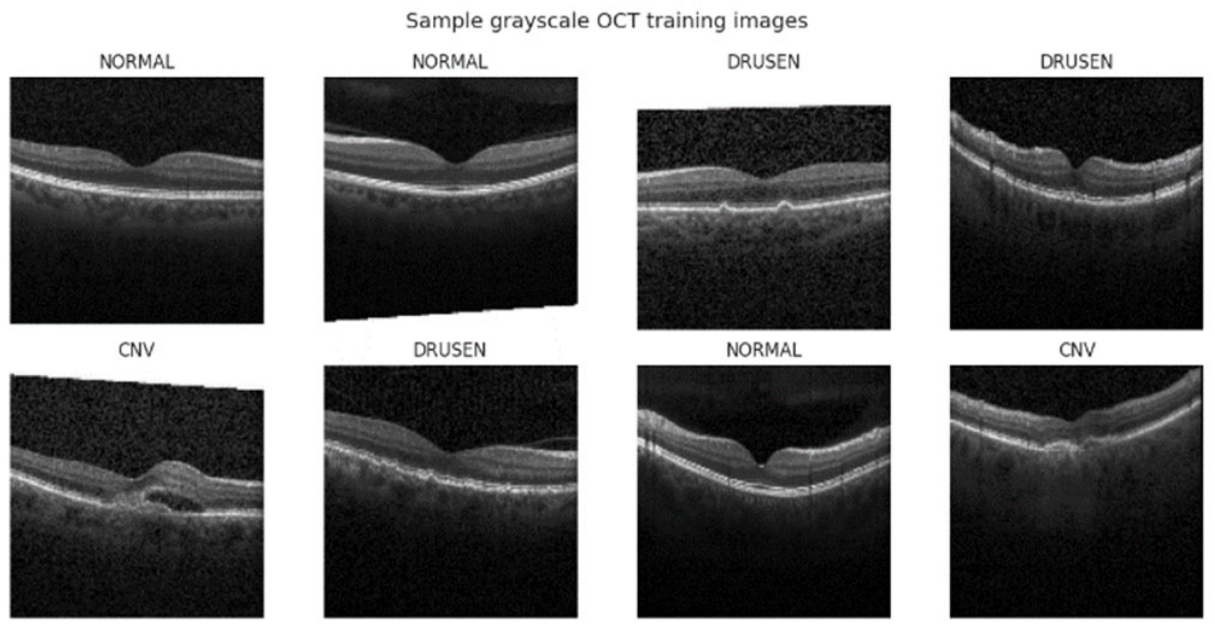

Fig. 5. Representative grayscale OCT images from the training set across the four diagnostic classes.

- **C.3. Augmentation rationale.** Horizontal flipping was used to introduce lateral variation while preserving the cross-sectional morphology of the retinal scan. Small rotations and translations were included to model modest acquisition and positioning variability. Limited zoom was used to introduce scale variation, while contrast perturbation was used to reduce sensitivity to modest intensity variation. Vertical flipping was excluded because it would invert anatomical orientation. All geometric transformations used nearest-neighbor filling, and stochastic augmentation was disabled during validation and inference.

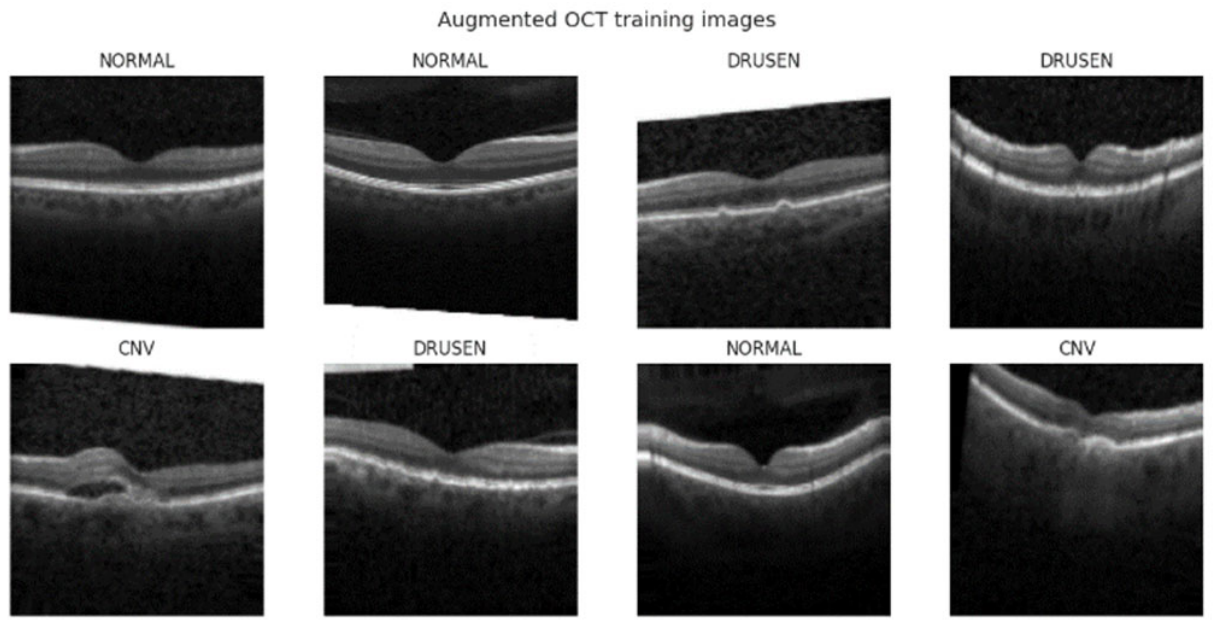

Fig. 6. Representative training images after stochastic data augmentation.

## D. Architecture

- ***D.1. Overall architecture.*** ED126-A is a purpose-built convolutional neural network designed for four-class retinal OCT classification. The model receives a $224 \times 224 \times 1$ grayscale input and consists of four convolutional blocks followed by a global average-pooling-based classification head. Each convolutional block contains two $3 \times 3$ convolutions followed by Batch Normalization and ReLU activation, with $2 \times 2$ stride-2 max pooling used for progressive spatial reduction. The final convolutional block is followed by spatial dropout, global average pooling, a 256-unit fully connected layer with ReLU activation, dense dropout, and a four-unit softmax output layer.

The architecture contains 1,241,380 total parameters, of which 1,239,460 are trainable. The network follows a hierarchical design paradigm, organized into six functionally distinct stages: an input preprocessing pipeline, four progressively deepening convolutional blocks, and a classification head. This staged organization is intended to encourage a hierarchical representation in which early layers encode local intensity and structural patterns, while deeper layers can represent increasingly abstract task-relevant features.

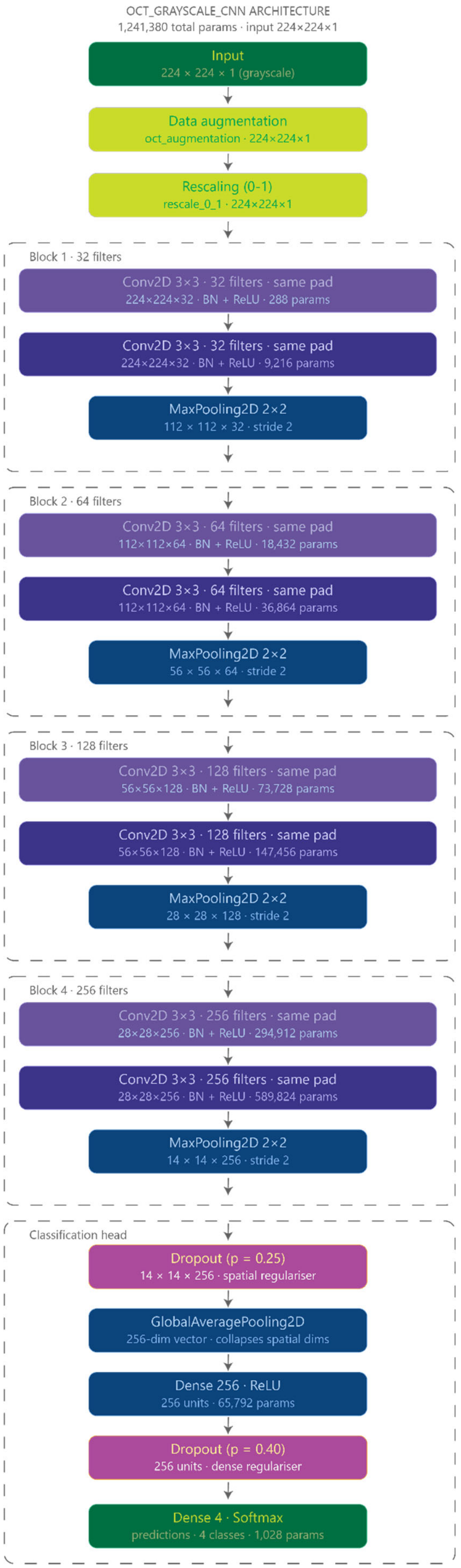

Fig. 7. Overall architecture of the ED126-A convolutional neural network.

The architecture is deliberately restrained in parameter count relative to ImageNet-scale models such as VGG-16 (~138M parameters) or ResNet-50 (~25M parameters), reflecting the data-efficiency demands of medical imaging tasks where labeled datasets are often substantially smaller in scale than the datasets used to train ImageNet-scale models. The filter schedule increases channel capacity from 32 to 256 across the four convolutional blocks while maintaining a compact overall parameter budget. Batch Normalization (BN) is applied after every convolutional layer, and no bias terms are used in convolutional layers. Regularization is addressed through a two-tier Dropout strategy and GlobalAveragePooling2D in the classification head, trading spatial resolution for resistance to overfitting.

- ***D.2. Convolutional blocks.***

The convolutional feature extractor comprises four sequential convolutional blocks with channel widths of 32, 64, 128, and 256 filters, respectively. Each block contains two $3 \times 3$ convolutional layers with same padding, followed by Batch Normalization (BN) and ReLU activation. No bias terms are used in the convolutional layers. A $2 \times 2$ max-pooling operation with stride 2 follows each block, producing progressive spatial reduction from the $224 \times 224$ input resolution to $112 \times 112$, $56 \times 56$, $28 \times 28$, and finally $14 \times 14$ feature maps.

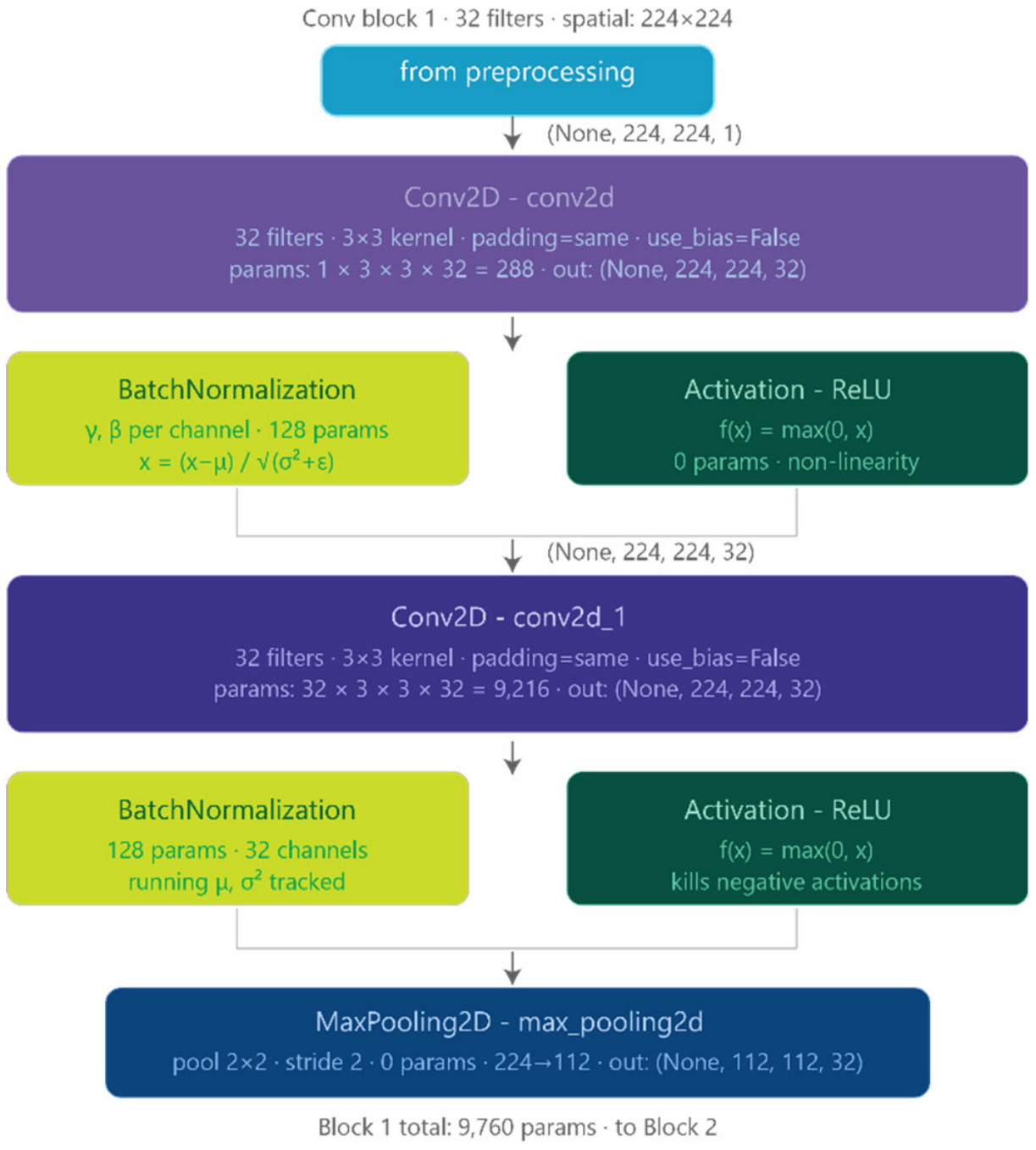

Fig. 8. Convolutional Block 1: 32 filters; spatial resolution reduced from $224 \times 224$ to $112 \times 112$.

**Block 1** operates on the $224 \times 224 \times 1$ input and uses 32 filters in both convolutional layers. The first convolution contains 288 kernel parameters, while the second contains 9,216, giving a total of 9,760 parameters for the block, including its Batch Normalization parameters. The block preserves the spatial resolution through same-padded convolutions and reduces it to $112 \times 112$ through max pooling. At this stage, the learned filters operate on the original grayscale representation and provide the initial transformation of local intensity and structural patterns.

**Block 2** increases the channel width from 32 to 64 filters while operating at 112 $\times$ 112 spatial resolution before pooling. Its two convolutional layers contain 18,432 and 36,864 kernel parameters, respectively, with Batch Normalization applied after each convolution. The block contains 55,808 parameters in total and produces a 56 $\times$ 56 $\times$ 64 output after max pooling. The increased channel dimension provides a larger feature space for representing patterns derived from the preceding block.

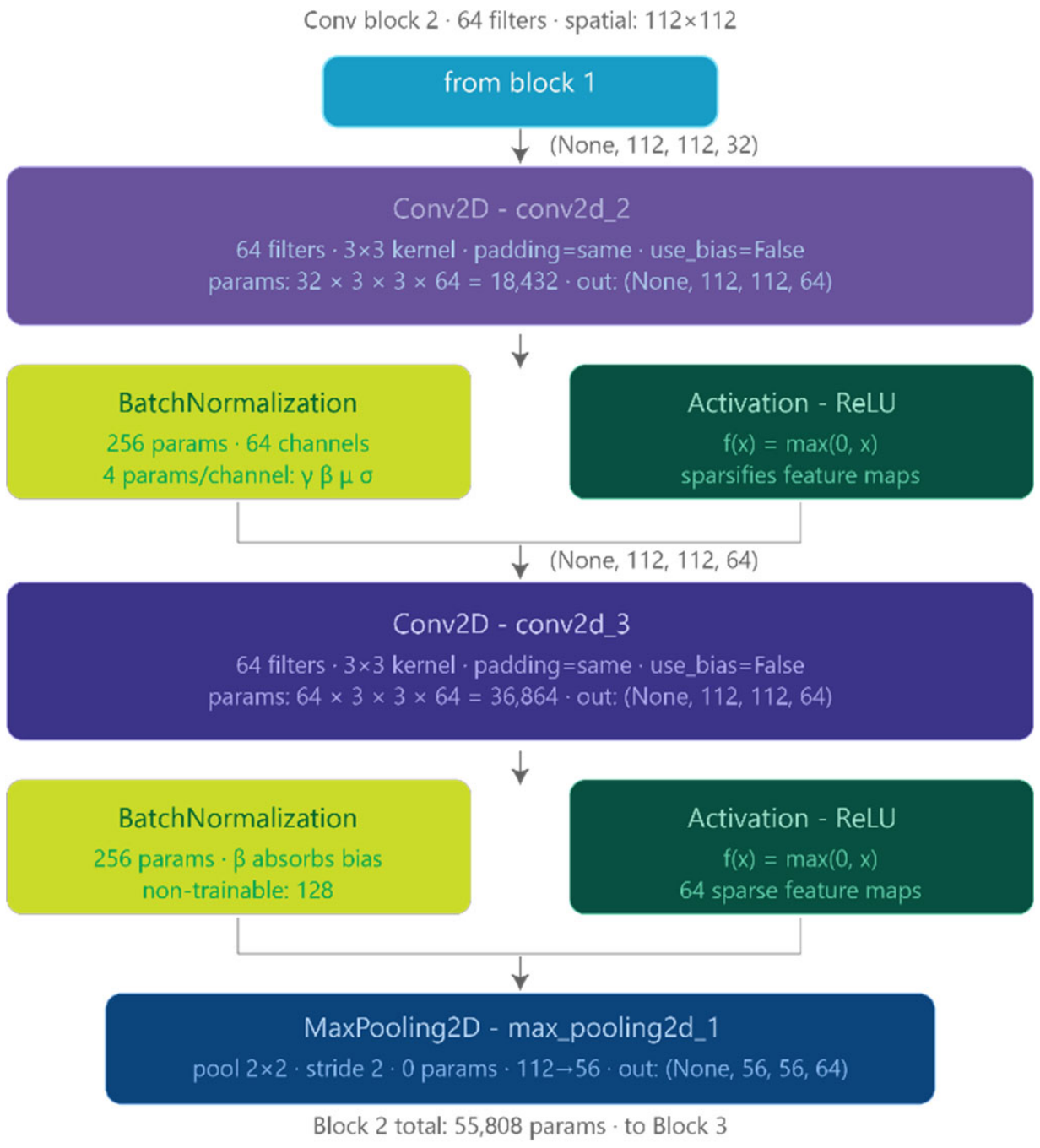

Fig. 9. Convolutional Block 2: 64 filters; spatial resolution reduced from 112 $\times$ 112 to 56 $\times$ 56.

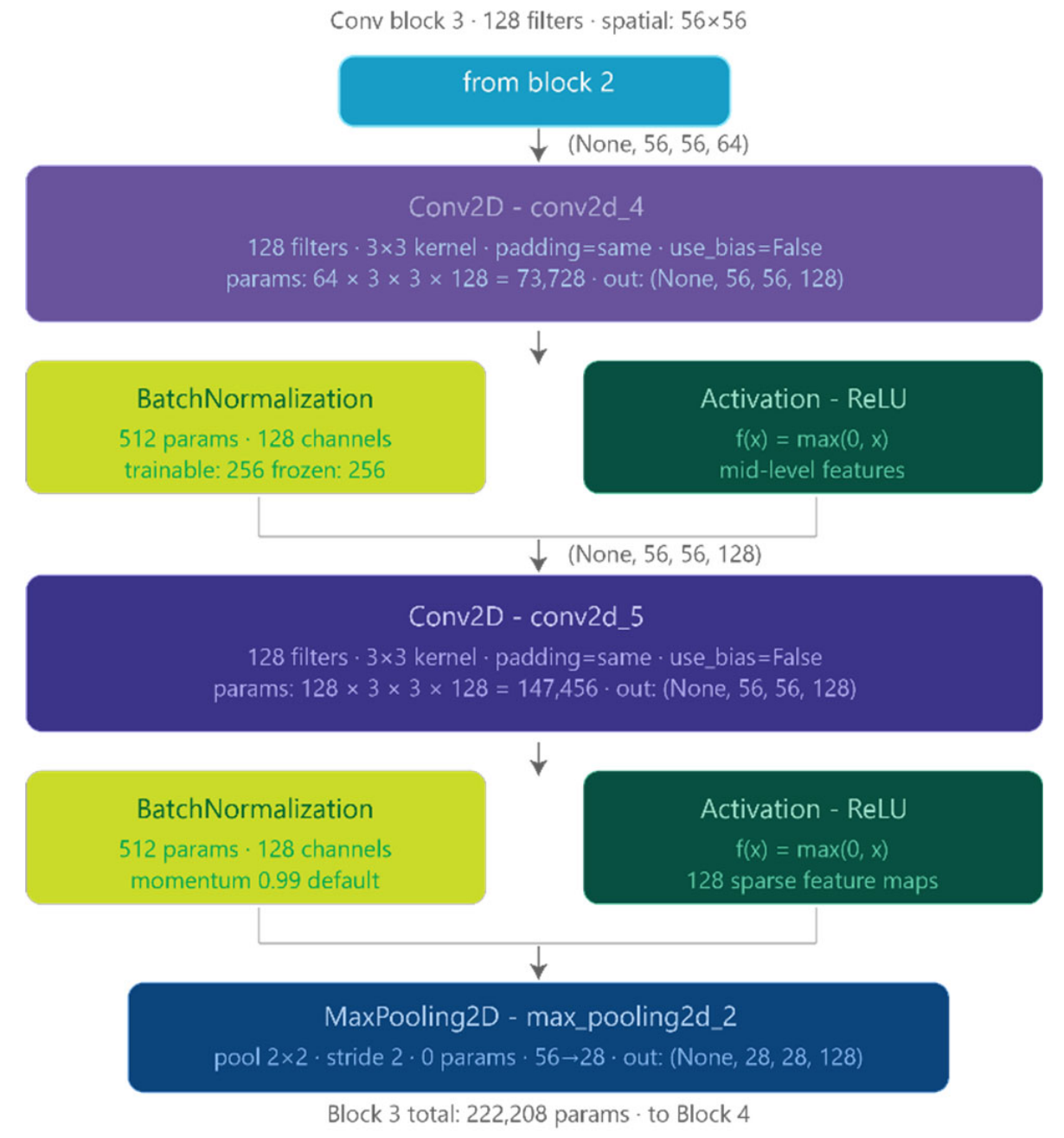

Fig. 10. Convolutional Block 3: 128 filters; spatial resolution reduced from 56 $\times$ 56 to 28 $\times$ 28.

**Block 3** further increases the channel width from 64 to 128 filters. Operating at 56 $\times$ 56 spatial resolution before pooling, its two convolutional layers contain 73,728 and 147,456 kernel parameters, respectively, resulting in 222,208 parameters for the complete block. Following the second BN-ReLU stage, 2 $\times$ 2 stride-2 max pooling reduces the spatial dimensions to 28 $\times$ 28 , producing an output tensor of 28 $\times$ 28 $\times$ 128 . The increasing channel capacity and reduced spatial resolution provide a progressively broader hierarchical feature representation.

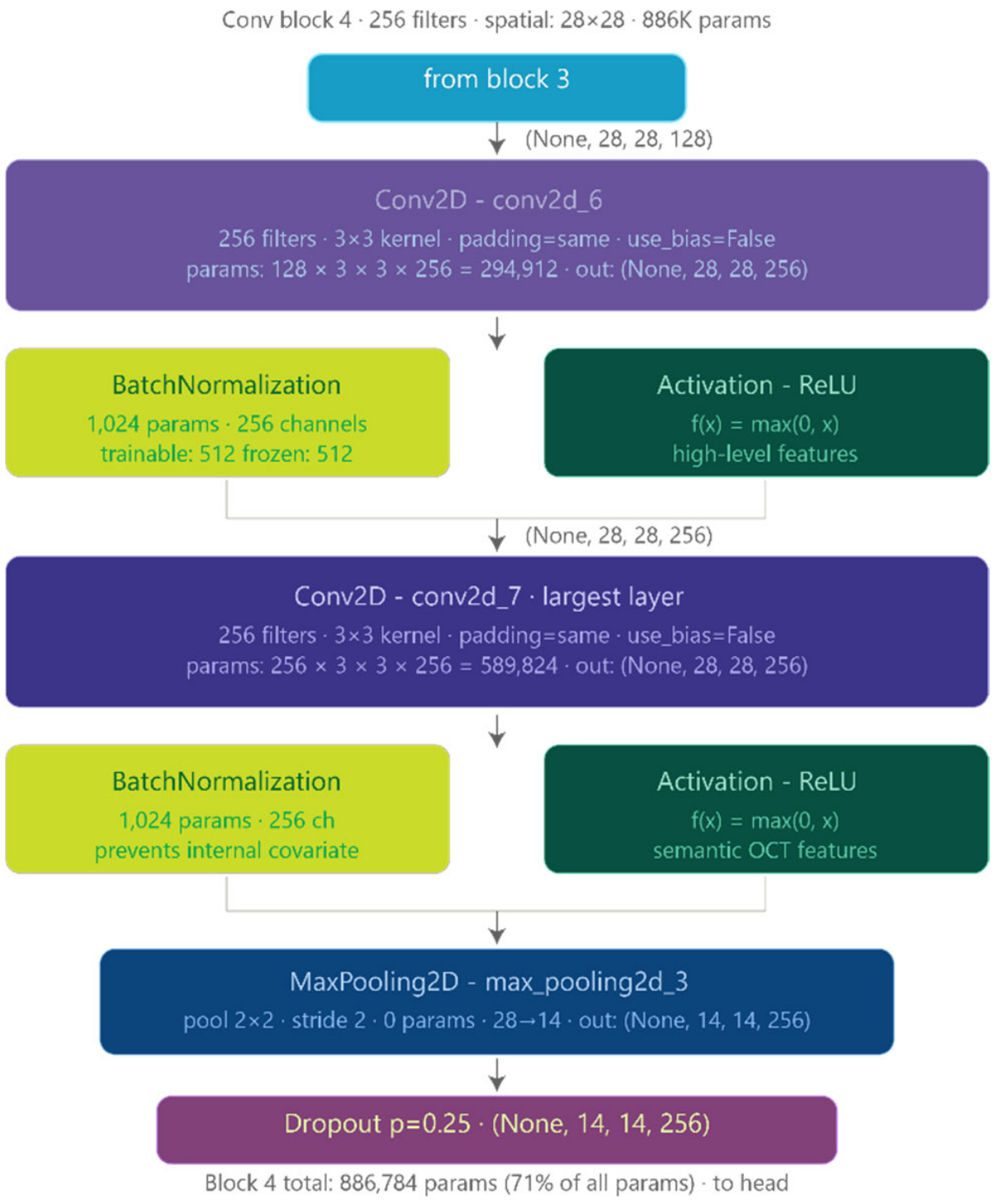

Fig. 11. Convolutional Block 4: 256 filters; spatial resolution reduced from 28 $\times$ 28 to 14 $\times$ 14.

**Block 4** is the deepest convolutional stage and increases the channel width to 256 filters. It operates at 28 $\times$ 28 spatial resolution before the final pooling operation. The first and second convolutional layers contain 294,912 and 589,824 kernel parameters, respectively, giving the block a total of 886,784 parameters, corresponding to 71.4% of the complete model parameter count. The second convolution is the largest parameterized layer in ED126-A. Following the final BN-ReLU operation, 2 $\times$ 2 stride-2 max pooling produces a 14 $\times$ 14 $\times$ 256 feature tensor.

A spatial dropout layer with a rate of 0.25 is applied to the output of Block 4 before the classification head. This operation regularizes the feature representation by randomly suppressing complete feature channels during training. The resulting 14 $\times$ 14 $\times$ 256 tensor is subsequently passed to the classification head described in Section III-D.3.

- ***D.3. Classification Head.***

The classification head receives the 14 $\times$ 14 $\times$ 256 feature tensor produced by the final convolutional block and transforms it into a four-class probability distribution. It comprises spatial dropout, Global Average Pooling (GAP), a 256-unit fully connected layer with ReLU activation, a conventional dropout layer, and a four-unit softmax output layer. The complete classification head contains 66,820 parameters.

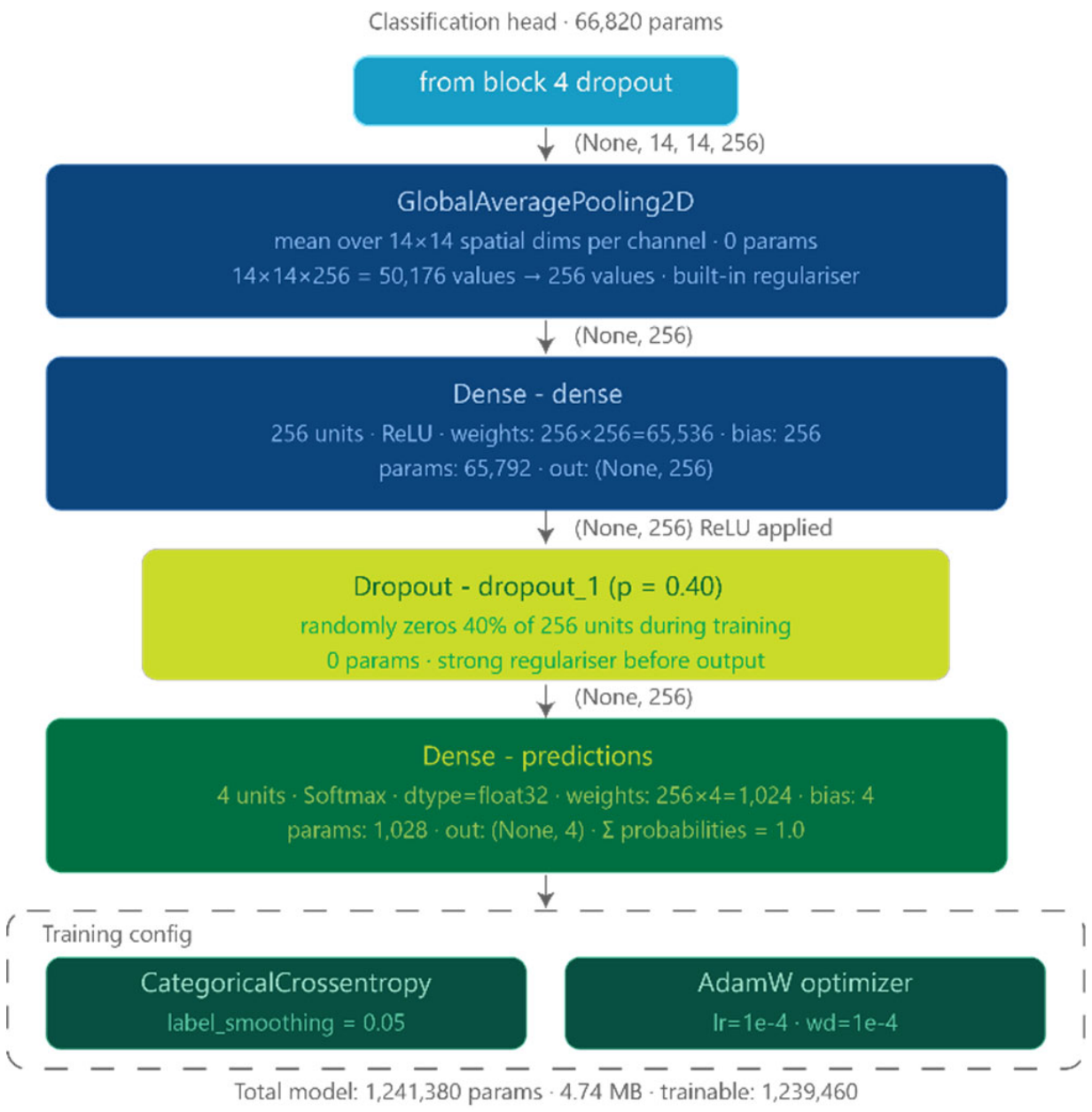

Fig. 12. Classification head of ED126-A comprising global average pooling, a fully connected layer, dropout, and a four-class softmax output.

*Spatial Dropout*. A spatial dropout rate of 0.25 is applied immediately after Block 4. During training, complete feature maps are randomly suppressed, introducing structured regularization at the end of the convolutional feature extractor.

*Global Average Pooling*. The $14 \times 14 \times 256$ tensor is reduced to a 256-dimensional vector by computing the spatial mean independently for each of the 256 channels. GAP contains no trainable parameters and avoids the substantially larger parameterization that would result from flattening all 50,176 spatial activations before the dense layer.

*Fully Connected Projection*. The resulting 256-dimensional vector is passed to a dense layer with 256 units and ReLU activation. This layer contains 65,792 parameters, including the weight matrix and bias vector, and provides the learned feature representation used by the final classifier.

*Dense Dropout*. A dropout rate of 0.40 is applied to the 256-dimensional dense representation during training before the output layer. The relative contribution of this regularization setting was not isolated experimentally in the present study.

*Output Layer*. The final dense layer contains four units with softmax activation and 1,028 parameters. It maps the 256-dimensional representation to four class logits and converts them into a probability distribution over CNV, DME, Drusen, and Normal, with the output probabilities summing to one. The classification head transforms the spatially-resolved tensors into a vector of class probabilities, employing a combination of global pooling, fully connected layers, and targeted regularization.

- ***D.4. Parameter Distribution***

TABLE III. PARAMETER DISTRIBUTION SUMMARY

| Parameter Distribution Summary | | |
|---|---|---|
| ***Stage*** | ***Parameters*** | ***% of Total*** |
| Preprocessing | 0 | 0.00% |
| Block 1 (32 filters) | 9,760 | 0.79% |
| Block 2 (64 filters) | 55,808 | 4.50% |
| Block 3 (128 filters) | 222,208 | 17.90% |
| Block 4 (256 filters) | 886,784 | 71.40% |
| Classification Head | 66,820 | 5.38% |
| Total | 1,241,380 | 100.00% |

Table III summarizes the distribution of model parameters across the principal architectural stages. The largest allocation occurs in Block 4, which contains 886,784 parameters, corresponding to 71.4% of the total model parameter count. This distribution reflects the selected channel progression from 32 to 256 filters and the increasing parameter cost of the deeper convolutional layers.

- ***D.5. Architectural rationale***

The ED126-A architecture was designed to provide a compact convolutional baseline while preserving a progressively hierarchical feature-extraction pathway for four-class retinal OCT classification. The channel configuration increases from 32 to 64, 128, and 256 filters across the four convolutional blocks, providing progressively greater representational capacity at deeper stages while maintaining a relatively small overall parameter count of 1,241,380. Spatial dimensions are reduced by $2 \times 2$ max pooling after each convolutional block, decreasing the feature-map resolution from $224 \times 224$ at the input to $14 \times 14$ after Block 4 and thereby reducing the computational cost of subsequent processing.

*Batch Normalization* is applied after each convolutional layer to standardize intermediate activations and is combined with ReLU nonlinearities. Regularization is provided through spatial dropout after the final convolutional block and conventional dropout in the fully connected classification head. The latter uses a higher dropout rate (0.40) than the spatial dropout stage (0.25), while the relative effect of these regularization choices is not isolated experimentally in the present study.

*Global Average Pooling (GAP)* converts the final $14 \times 14 \times 256$ feature tensor into a 256-dimensional representation by aggregating each channel spatially. This design avoids the much larger parameterization that would result from flattening the complete spatial tensor before the dense layer. The resulting representation is passed to a 256-unit dense layer and a four-unit softmax output corresponding to CNV, DME, Drusen, and Normal.

Overall, the architecture deliberately allocates most of its parameters to the deepest convolutional stage, where the channel count reaches 256, while maintaining comparatively lightweight earlier stages and a compact classification head. This configuration defines the architectural baseline evaluated in the present study and provides a fixed reference condition for subsequent experiments involving transfer learning, increased depth, attention, or multi-scale feature fusion.

- **E. Training Configuration and Class-Imbalance Handling**

- ***E.1. Optimization and regularization.***

ED126-A was optimized using categorical cross-entropy with label smoothing of 0.05 and the AdamW optimizer with an initial learning rate of $1 \times 10^{-4}$ and weight decay of $1 \times 10^{-4}$. Training used batches of eight images for a maximum of 30 epochs. The learning rate was reduced by a factor of 0.5 after two epochs without improvement in validation loss, with a minimum learning rate of $1 \times 10^{-6}$. Early stopping with a patience of four epochs was used, with the best validation-loss checkpoint restored. Mixed-precision training was enabled, and all experiments used random seed 42.

***Label smoothing***. A smoothing coefficient of $\epsilon = 0.05$ was used to soften the target distribution and reduce the incentive for excessively confident predictions.

TABLE IV. TRAINING CONFIGURATION

| ***Training Configuration*** | |
|---|---|
| ***Parameter*** | ***Value*** |
| Learning rate | 1e-4 |
| Weight decay | 1e-4 |
| Batch size | 8 |
| Number of epochs | 30 (max) |
| Learning rate schedule | ReduceLROnPlateau (factor=0.5, patience=2, min_lr=1e-6, monitor=val_loss) |
| Early stopping | Yes - patience=4, monitor=val_loss, restore_best_weights=True |
| Best checkpoint selection | Best model on val_loss (ModelCheckpoint, save_best_only=True) |
| Mixed precision | Mixed_float16 policy |
| Conv kernel size | 3 × 3 |
| All padding | Same |
| All pooling | 2 × 2, stride 2 |
| BN momentum | 0.99 |
| Label smoothing | 0.05 |
| Spatial dropout (Block 4) | 0.25 |
| Dense dropout (head) | 0.40 |
| Class weighting | Yes - balanced class weights (scikit-learn formula: total / (n_classes * count_i)) |
| Seed | 42 |
| Training GPU | NVIDIA T4 (16 GB VRAM) |
| Inference GPU | NVIDIA GTX 960M (4 GB VRAM) |

- ***E.2. Class Imbalance handling***: Because the training distribution is heterogeneous across classes, balanced class weights were applied during optimization using the standard inverse-frequency formulation:

$$w_i = N_{\text{total}}/(N_{\text{classes}} \times N_i)(1)$$

where $N_i$ denotes the number of training samples belonging to class $i$. This weighting increases the contribution of minority-class errors to the training objective. The persistence of the Drusen precision deficit under this configuration is reported as an empirical result rather than interpreted as evidence that class weighting is intrinsically insufficient.

- ***E.3. Reproducibility and computational environment***: All experiments were conducted using a fixed random seed of 42 and mixed-precision training. Training was performed on an NVIDIA T4 GPU with 16 GB VRAM, while inference was evaluated on an NVIDIA GTX 960M GPU with 4 GB VRAM. The reported results correspond to a single training run.

- **F. Evaluation Protocol**

ED126-A is subjected to a multi-faceted evaluation suite encompassing:

- **Training dynamics**: loss and accuracy curves across the full training trajectory, inspected for convergence stability, overfitting signatures, and learning rate scheduling events.
- **Class-wise performance**: sensitivity (recall), specificity, precision, and F1-score computed for each of the four diagnostic classes, complemented by macro-averaged and weighted aggregates.
- **Confusion matrix**: both absolute and row-normalized forms, enabling identification of systematic inter-class confusions.
- **Confidence analysis**: top-2 softmax probabilities and decision margins.
- **Representation analysis**: t-SNE projection of the penultimate-layer feature embedding to visualize qualitative patterns of class clustering and local overlap in the learned representation.
- **Interpretability**: Gradient-weighted Class Activation Mapping (Grad-CAM [23]) applied to correctly and incorrectly classified samples to localize discriminative regions and diagnose failure modes.
- **Hierarchical feature analysis**: intermediate feature map visualization across selected convolutional stages to interrogate the progression of learned representations from low-level edge detection to task-relevant pathological features.
- **Threshold-independent discrimination**: per-class Receiver Operating Characteristic (ROC) curves with corresponding Area Under the Curve (AUC) scores, computed under a one-vs-rest strategy to quantify each class's discriminative capacity across the full decision threshold spectrum, complemented by Precision-Recall (PR) curves and their associated Average Precision (AP) scores.

## IV. RESULTS

All results reported in this section were obtained from ED126-A evaluated on the 10,933-image image-level held-out test set. No transfer learning, architectural enrichment, or model ensembling was used. The model was trained once using the experimental configuration described in Section III, and the final evaluation was performed using the checkpoint selected by minimum validation loss at epoch 27.

### A. Training Dynamics

ED126-A was trained for a maximum of 30 epochs using an initial learning rate of $1 \times 10^{-4}$, a ReduceLROnPlateau schedule with a reduction factor of 0.5 and patience of two epochs, and early stopping with a patience of four epochs. The model was optimized end-to-end from random initialization without staged freezing or fine-tuning, and mixed-precision training was enabled throughout. Because the experiment comprised a single training run, all reported performance values should be interpreted as point estimates for the specified initialization and training configuration. The best checkpoint was obtained at epoch 27, with a validation loss of 0.2953, validation accuracy of 0.9611, and training accuracy of 0.9766.

TABLE V. SELECTED EPOCH TRAINING METRICS

| *Epoch* | *LR* | *Train Loss* | *Train Acc* | *Val Loss* | *Val Acc* |
|---|---|---|---|---|---|
| 0 | 1.00e-4 | 0.7301 | 0.7797 | 0.6522 | 0.8234 |

| Ep och | LR | Train Loss | Train Acc | Val Loss | Val Acc |
|---|---|---|---|---|---|
| 5 | 1.00e-4 | 0.3599 | 0.9413 | 0.4966 | 0.8775 |
| 10 | 1.00e-4 | 0.3277 | 0.9545 | 0.3742 | 0.9246 |
| 11 | 5.00e-5 | 0.3088 | 0.9611 | 0.3313 | 0.9420 |
| 15 | 2.50e-5 | 0.2897 | 0.9679 | 0.3146 | 0.9511 |
| 20 | 1.25e-5 | 0.2774 | 0.9724 | 0.2995 | 0.9591 |
| 25 | 6.25e-6 | 0.2699 | 0.9751 | 0.3061 | 0.9555 |
| 26 | 6.25e-6 | 0.2675 | 0.9762 | 0.3155 | 0.9513 |
| 27 | 6.25e-6 | 0.2670 | 0.9766 | 0.2953 | 0.9611 |
| 29 | 6.25e-6 | 0.2651 | 0.9778 | 0.3103 | 0.9541 |

Figure 13 shows the training and validation trajectories across the 30-epoch training process. Training accuracy increased from 0.7797 at epoch 0 to 0.9778 at epoch 29. Validation accuracy increased overall but exhibited greater variation during the early and intermediate stages of optimization, ranging from approximately 0.7782 to 0.9484 during epochs 0-10. Following the first learning-rate reduction at epoch 11, the magnitude of the observed fluctuations decreased, and validation performance became progressively more stable.

At epoch 1, validation accuracy temporarily fell below training accuracy, producing a short-lived generalization discrepancy. After the first learning-rate reduction at epoch 11, the validation trajectory showed attenuated oscillations, while both training and validation accuracy approached a stable region during the later epochs. At epoch 29, the difference between training and validation accuracy was 0.0237, corresponding to training accuracy of 0.9778 and validation accuracy of 0.9541. Validation loss remained in the vicinity of 0.30 during the later stages of training, with the minimum value of 0.2953 occurring at epoch 27.

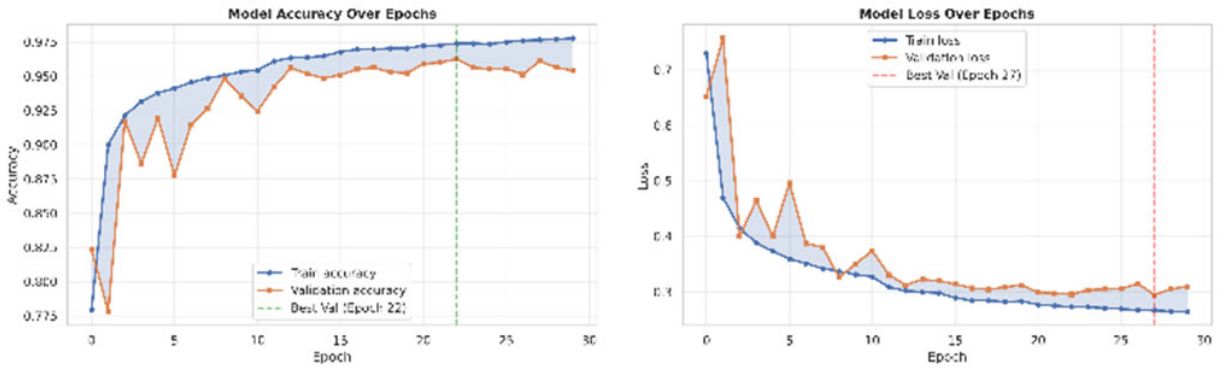

Fig. 13. Training and validation loss and accuracy trajectories across the 30-epoch training process.

## B. Per-Class Discriminative Metrics

The final performance of ED126-A on the 10,933-image image-level held-out test set is summarized in Table VI. All metrics were computed using the best-validation-loss checkpoint at epoch 27, without threshold tuning or post-hoc calibration.

TABLE VI. PER-CLASS PERFORMANCE METRICS

| Nr | class | Sensiti vity | Specifi city | precisi on | recall | f1_scor e |
|---|---|---|---|---|---|---|
| 0 | CNV | 0.9285 | 0.9949 | 0.9895 | 0.9285 | 0.9580 |
| 1 | DME | 0.9483 | 0.9824 | 0.8649 | 0.9483 | 0.9047 |
| 2 | DRUSEN | 0.9605 | 0.9716 | 0.7493 | 0.9605 | 0.8419 |
| 3 | NORMAL | 0.9650 | 0.9915 | 0.9902 | 0.9650 | 0.9774 |
| | accuracy | | | 0.9503 | 0.9503 | 0.9503 |
| | macro avg | | | 0.8985 | 0.9506 | 0.9205 |
| | weighted avg | | | 0.9571 | 0.9503 | 0.9521 |

ED126-A achieved an overall accuracy of 95.03%, with a macro-F1 of 0.9205 and a weighted-F1 of 0.9521. The difference between weighted and macro F1 reflects the influence of the heterogeneous class distribution on aggregate performance measures.

- **Normal class**: achieved the highest F1-score at 0.9774, with precision of 0.9902 and sensitivity of 0.9650. The corresponding results indicate that false-positive Normal predictions were infrequent at the reported operating point, while 3.5% of true Normal samples were assigned to another class. The dominant error direction for these misclassified Normal samples was the Drusen class.

- **CNV class**: achieved an F1-score of 0.9580, with precision of 0.9895 and sensitivity of 0.9285. Its specificity was 0.9949, indicating a low frequency of false-positive CNV predictions. Among the 268 misclassified CNV samples, 176 were assigned to Drusen, making CNV → Drusen the largest individual CNV error channel.

- **DME class**: achieved an F1-score of 0.9047, with sensitivity of 0.9483 and precision of 0.8649. Thus, DME sensitivity exceeded precision at the reported operating point. Of the 60 DME errors, 33 were assigned to Normal and 14 to CNV, making DME → Normal the largest DME error channel.

- **Drusen class** exhibited the lowest F1-score at 0.8419. Its sensitivity of 0.9605 was the highest among the four classes, whereas its precision of 0.7493 was the lowest. Consequently, the principal class-conditional limitation was not failure to identify true Drusen samples, but the assignment of the Drusen label to samples belonging to other classes. Among the 1,137 test predictions assigned to Drusen, 285 were false positives, corresponding to 25.1% of Drusen-positive predictions.

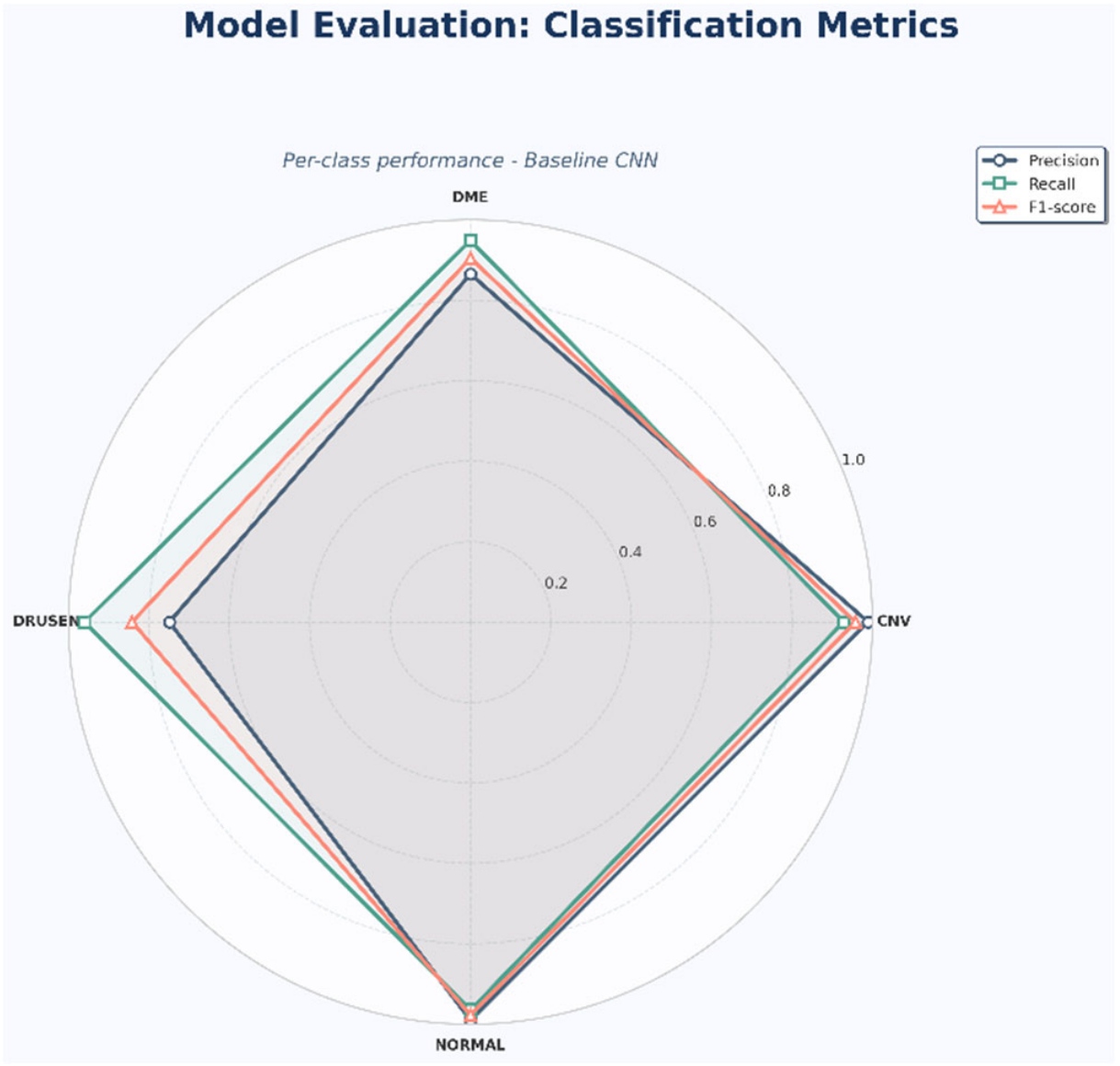

Fig. 14. Class-wise precision, recall, and F1-score for ED126-A on the image-level held-out test set.

## C. Confusion Matrix Analysis

The confusion matrix provides a class-level decomposition of the 543 test-set errors. Table VII reports the absolute prediction counts, while Figure 15 presents the corresponding row-normalized confusion matrix.

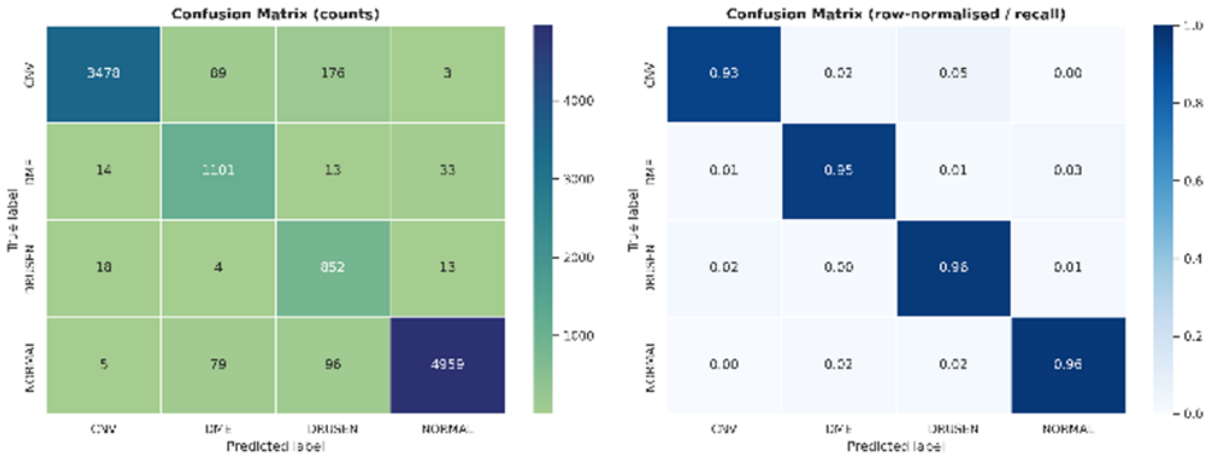

Fig. 15. Confusion matrix of ED126-A predictions on the image-level held-out test set.

TABLE VII. CONFUSION MATRIX

| ***True / Predicted*** | ***CNV*** | ***DME*** | ***DRUSEN*** | ***NORMAL*** | ***Total*** |
|---|---|---|---|---|---|
| ***CNV*** | 3478 | 89 | 176 | 3 | 3746 |
| ***DME*** | 14 | 1101 | 13 | 33 | 1161 |
| ***DRUSEN*** | 18 | 4 | 852 | 13 | 887 |
| ***NORMAL*** | 5 | 79 | 96 | 4959 | 5139 |

Key findings from the confusion structure:

- **Normal and CNV diagonal dominance**: The Normal and CNV rows exhibit strong diagonal dominance, consistent with sensitivities of 0.9650 and 0.9285 and specificities of 0.9915 and 0.9949, respectively. The principal off-diagonal errors are concentrated in a small number of class-pair channels.
- **CNV to Drusen dominant confusion**: The largest individual off-diagonal count is the CNV → Drusen channel, with 176 errors. These account for 4.70% of all CNV samples and 65.7% of the 268 CNV errors. CNV → DME represents the second-largest CNV error channel, with 89 cases, corresponding to 2.38% of the CNV class.
- **Normal-to-Drusen confusion and Drusen false-positive assignments**: The Normal → Drusen channel contains 96 errors, corresponding to 1.87% of the Normal class and 53.3% of its 180 errors. The Drusen prediction column receives false-positive assignments from all three alternative classes: 96 from Normal, 176 from CNV, and 13 from DME. Thus, 285 of the 1,137 Drusen-positive predictions are false positives, with CNV representing the largest individual source of these false-positive assignments.
- **DME error profile**: The DME class produced 60 errors in total, with 33 assigned to Normal and 14 assigned to CNV. These two channels account for 55.0% and 23.3% of all DME errors, respectively.

## D. Softmax Score and Decision-Margin Analysis

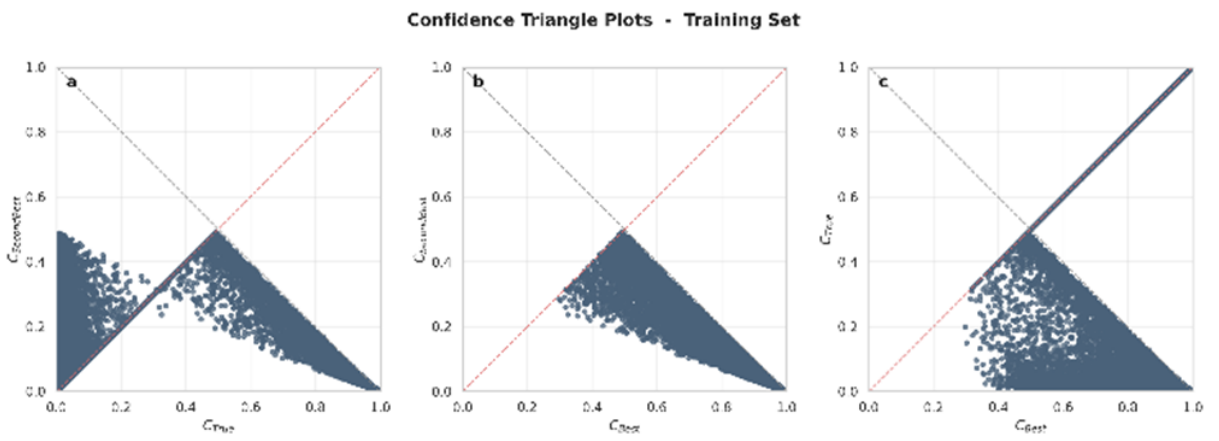

Fig. 16. Top-two softmax confidence structure for the training set, showing true-class or best-class probability against second-best-class probability.

To characterize prediction behavior beyond the discrete class labels, the top-two softmax probabilities were examined across the training, validation, and test partitions. Three complementary projections were considered: (a) true-class probability versus competing-class probability, (b) highest predicted probability versus winner-runner-up margin, and (c) highest predicted probability versus true-class probability. Figures 16-18 present these analyses for the training, validation, and test sets, respectively.

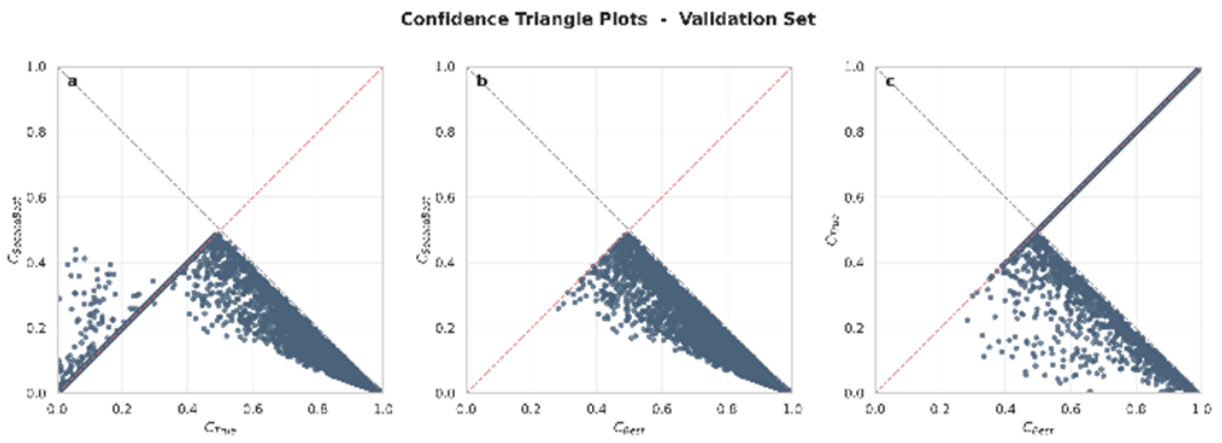

Fig. 17. Top-two softmax confidence structure for the validation set, showing true-class or best-class probability against second-best-class probability.

### D.1 True-Class versus Competing-Class Probability.

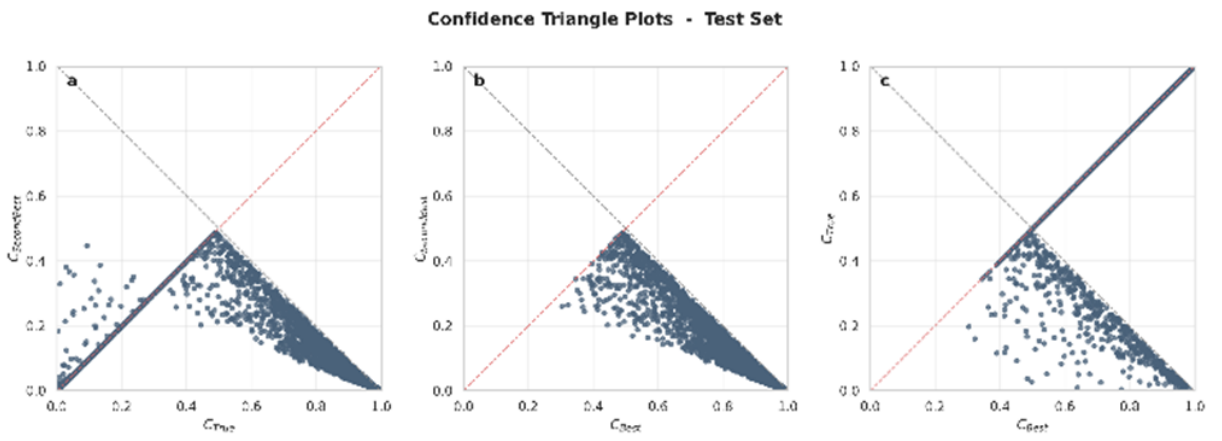

Fig. 18. Top-two softmax confidence structure for the image-level held-out test set, showing true-class or best-class probability against second-best-class probability.

The feasible region is defined by ($C_{\text{True}} + C_{\text{Second}} \leq 1$), with the diagonal ($C_{\text{True}} \approx C_{\text{Second}}$) representing greater pairwise ambiguity between the true and competing classes. Across all three partitions, a population of samples exhibited relatively low true-class probability while the strongest competing class retained a comparatively large probability mass. This pattern was particularly visible in the training set, where competing-class probabilities extended to approximately 0.5.

A broad diagonal band with both probabilities in the approximate range of 0.3-0.5 was also visible across the three partitions. The persistence of this pattern indicates recurring pairwise ambiguity in the model's predictions rather than a phenomenon isolated to a single evaluation split.

### D.2 Winner-Runner-Up Decision Margin.

The second projection characterizes the separation between the highest predicted probability and the second-highest probability. A prediction close to ( $C_{\text{Best}} = C_{\text{Second}}$ ) corresponds to a small winner-runner-up margin.

ED126-A produced a broad wedge-shaped distribution across approximately ($C_{\text{Best}} \in [0.4,0.7]$) and ($C_{\text{Second}} \in [0.2,0.45]$). The training set exhibited a particularly dense concentration near the diagonal region, whereas the validation and test distributions were comparatively attenuated. Across all three partitions, the absence of a dominant concentration at large winner-runner-up margins indicates that a substantial subset of predictions retained competing-class probability mass.

This pattern is consistent with ambiguity in the learned decision function. Because no explicit calibration analysis was performed, the relationship between these margins and predictive calibration cannot be determined from the present experiment.

### D.3 Highest Predicted Probability versus True-Class Probability.

The third projection separates correct predictions from errors based on the relationship between the highest predicted probability and the probability assigned to the true class. Correct top-1 predictions lie on the diagonal ($C_{\text{Best}} = C_{\text{True}}$), whereas misclassified samples occupy the off-diagonal region.

A dense diagonal population extended from approximately ($C_{\text{Best}} = 0.4$) to 0.98, corresponding to the dominant correctly classified population. At the same time, the off-diagonal error population extended into an intermediate-confidence region, including samples with ($C_{\text{Best}}$) approximately between 0.5 and 0.8 and ($C_{\text{True}}$) below approximately 0.3. Thus, a non-trivial subset of errors occurred with moderate-to-high predicted confidence rather than being restricted to low-confidence predictions.

The same qualitative off-diagonal morphology was visible across the evaluated partitions without substantial amplification from training to validation or test. This observation provides no strong evidence that the residual error pattern is driven exclusively by late-stage overfitting.

### E. t-SNE Representation-Space Analysis

The 256-dimensional embedding produced by the Global Average Pooling layer was projected into two dimensions using t-SNE with perplexity = 30 over 1,000 iterations and random seed 42. Figure 19 presents the projection for the image-level held-out test set using ground-truth class labels and prediction correctness.

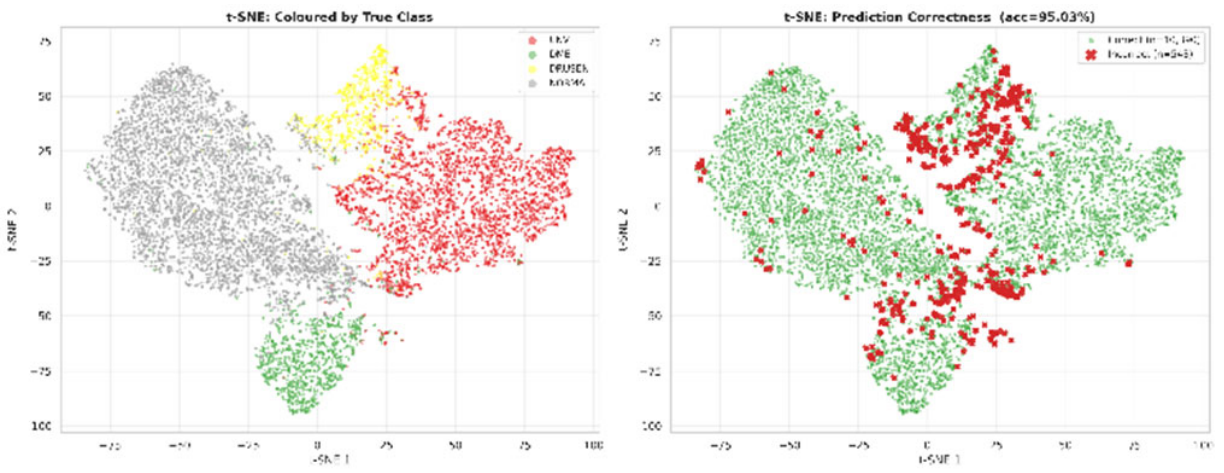

Fig. 19. t-SNE projection of the 256-dimensional GAP embeddings for the image-level held-out test set, shown by ground-truth class and prediction correctness.

- **Cluster loci and compactness**: A compact yellow cluster (top-center) corresponded to DRUSEN. A compact green cluster (bottom-left) corresponded to DME. A group of three red clusters in the middle-right quadrant, with moderate spread and mutual overlap, corresponded to CNV.

- **Inter-cluster overlap**: The DME-Normal boundary exhibited a broad transition zone, with DME samples showing greater spatial dispersion toward the Normal region. The CNV-DRUSEN boundary showed a tighter but multi-lobed interface, with the three red CNV subclusters interlocking at the yellow (DRUSEN) cluster's right flank rather than forming a single clean margin.

- **Transition-zone samples**: The diffuse light-green (Normal) cluster populated the central transition zone between the DME and CNV/Drusen regions, consistent with localized overlap in the projected embedding.

- **Misclassified sample distribution**: In the correctness-colored panel, misclassified samples (red) formed contiguous blocks concentrated in the boundary regions between the CNV/Drusen and Normal/DME/CNV loci rather than in cluster interiors. Correctly classified samples (green) occupied the cluster cores. The spatial distribution is consistent with misclassifications being concentrated near regions of local class overlap in the projected representation space.

- **Overall assessment**: The t-SNE visualization is qualitatively consistent with the class-conditional performance results: regions of greater apparent overlap contain a larger proportion of misclassified samples, while visually compact regions contain a greater concentration of correctly classified samples.

### F. Grad-CAM Interpretability

Grad-CAM saliency maps were generated by backpropagating the predicted-class score to the terminal convolutional layer, conv2d_8, containing 256 filters. Four correctly classified and four misclassified test instances were selected from the first ten test batches without class stratification. Because the test set is ordered by class folder, the sampled instances represented CNV only. Accordingly, this analysis is restricted to qualitative examples of CNV predictions and does not characterize Grad-CAM behavior for DME, Drusen, or Normal samples.

#### F.1 Correctly Classified CNV Examples

All four correctly classified examples had the ground-truth label CNV and were assigned CNV with predicted probabilities ranging from 0.891 to 0.961. The corresponding Grad-CAM maps were relatively focal, with activation concentrated in one or two localized regions. In the overlays, the high-activation regions were visually localized within retinal structures that appeared relevant to the classification decision.

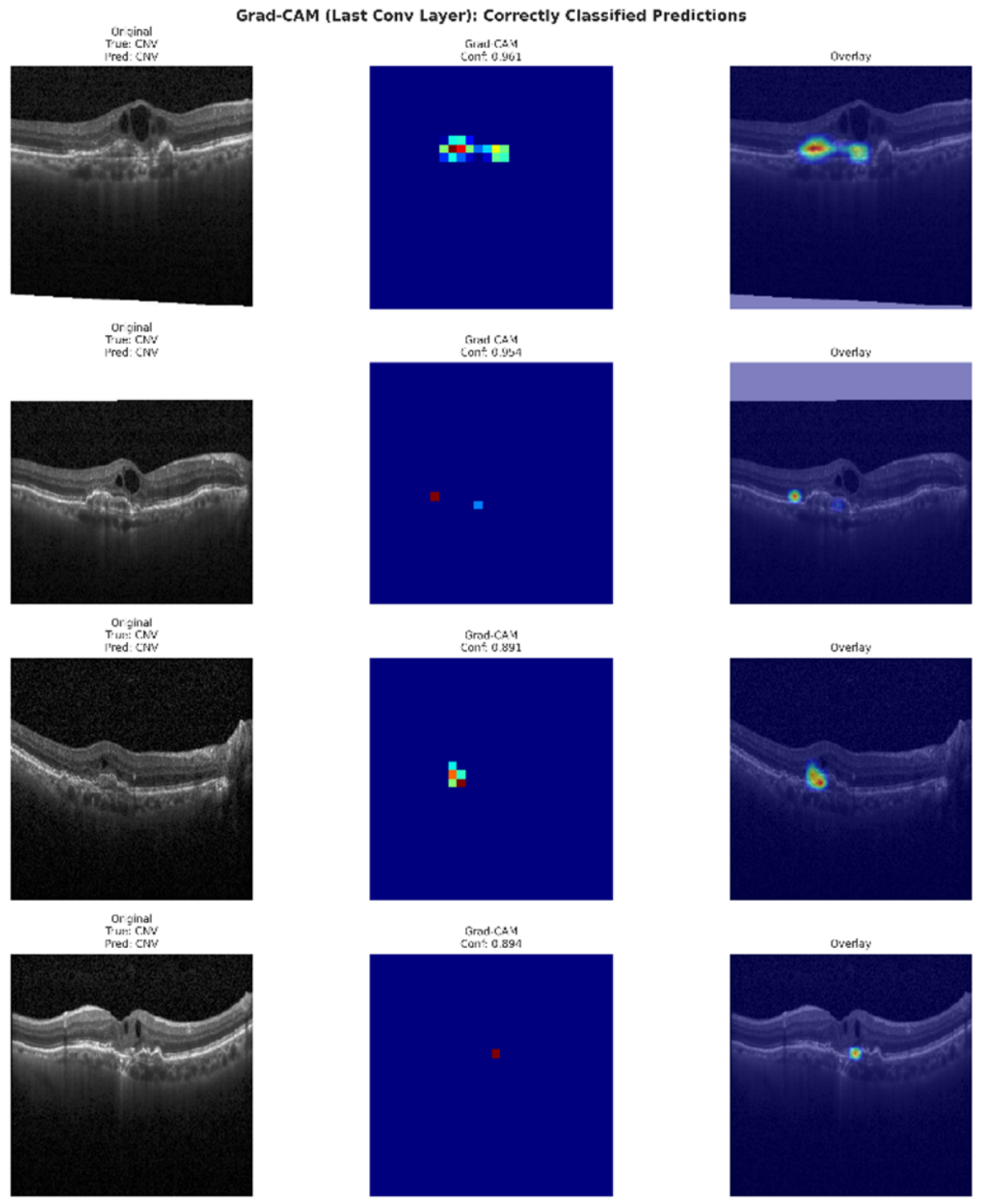

Fig. 20. Grad-CAM visualizations for correctly classified CNV test images.

### F.2 Misclassified CNV Examples

All four misclassified examples also had the ground-truth label CNV. Three were predicted as DME, with predicted probabilities of 0.938, 0.732, and 0.568, while one was predicted as Drusen with a predicted probability of 0.541. Compared with the correctly classified examples, the corresponding Grad-CAM maps appeared less focal, with elongated, multi-lobed, or spatially separated activation patterns.

These visual differences are consistent with less spatially concentrated model activation in the selected misclassified examples. Because activation location was not quantitatively measured, the analysis should be interpreted as qualitative rather than as evidence of a statistically established shift in localization.

The CNV-only sampling limits the scope of this analysis. In particular, the presented examples cannot be used to infer Grad-CAM behavior for Drusen, DME, or Normal classes or to quantify class-pair-specific interpretability patterns.

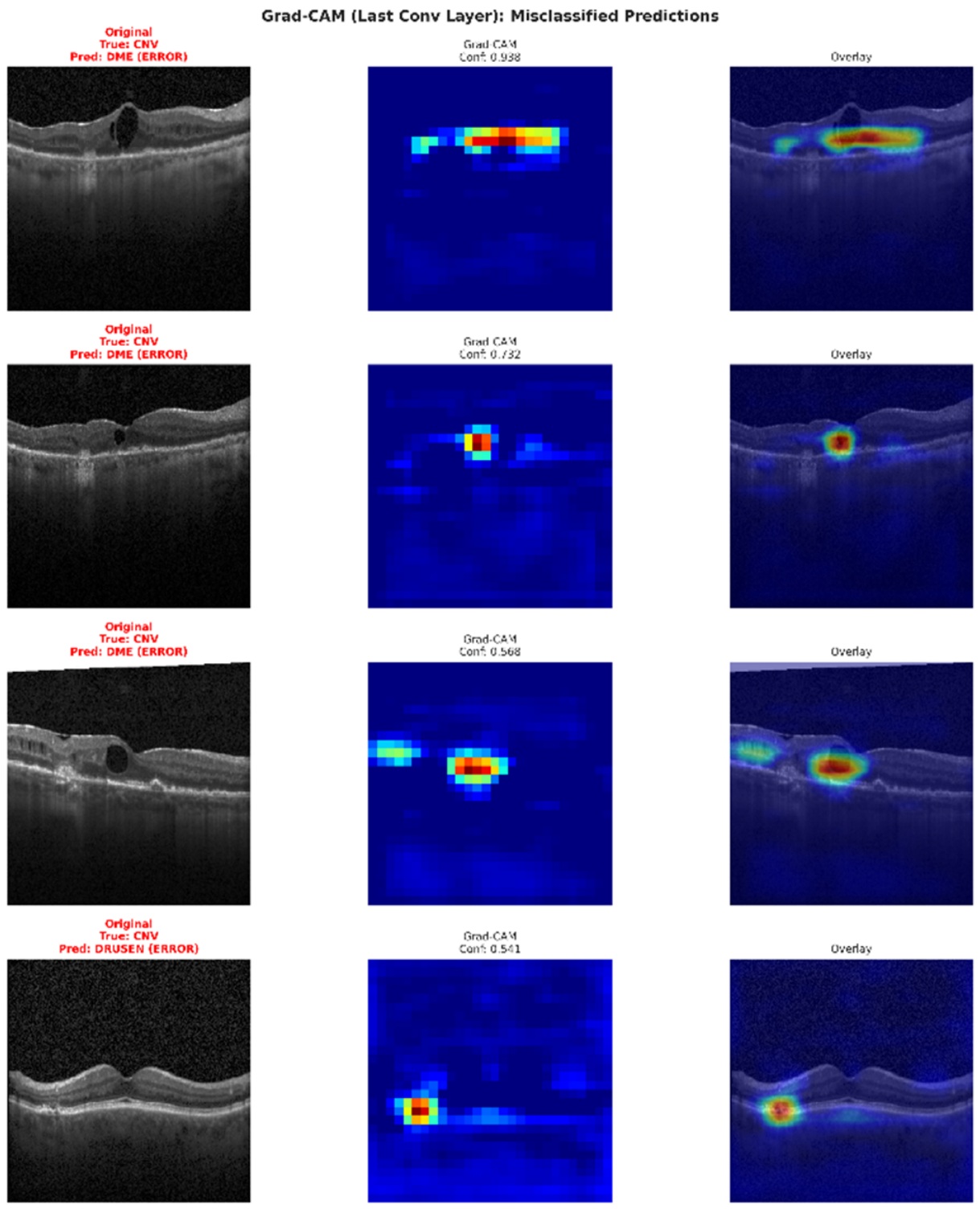

Fig. 21. Grad-CAM visualizations for misclassified CNV test images.

## G. Hierarchical Feature Map Analysis

Intermediate activations were extracted from the eight convolutional layers of the trained model using a functional sub-model. Sixteen filters were visualized from each layer, producing 128 displayed feature maps in an $8 \times 16$ visualization grid. No uniformly inactive filters were observed among the 128 displayed maps; all displayed filters exhibited visible structural variation.

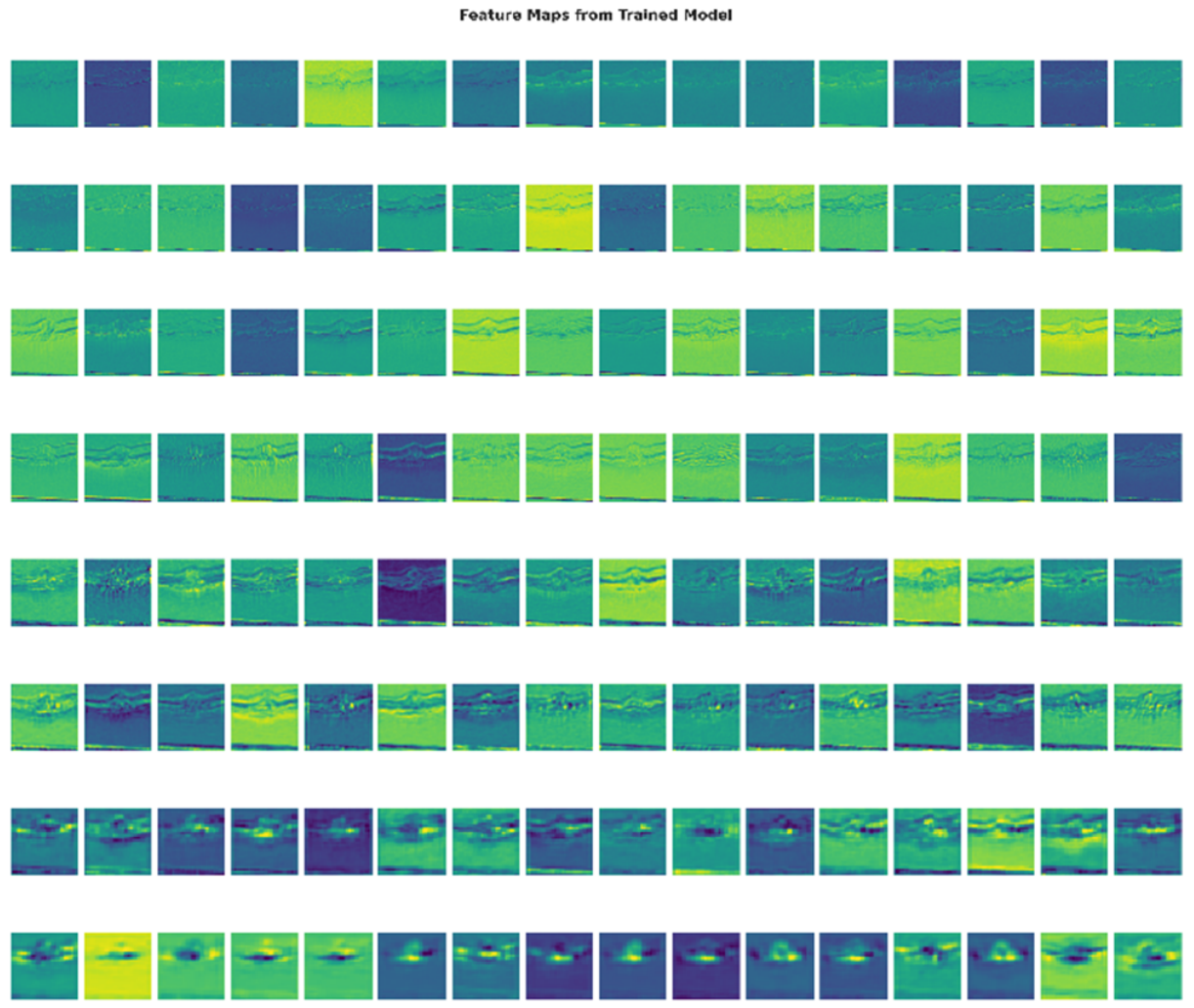

Fig. 22. Representative feature-map activations extracted from the eight convolutional layers of ED126-A.

**Layer-wise progression**: The shallow convolutional layers predominantly produced edge- and texture-like activation patterns, including oriented structures, luminance gradients, and retinal boundary responses. At intermediate depths, activations exhibited more curved, localized, and blob-like structures. Deeper layers showed increasingly localized activation patterns that visually corresponded to retinal structures and regions associated with the classification task.

**Hierarchical summary**: The observed progression can be summarized as follows: Layers 1-2 predominantly represented edges and retinal-layer boundaries; Layers 3-4 emphasized textures, contours, and ovoid structures; Layers 5-6 produced more localized structural responses; and Layers 7-8 exhibited sparse, high-level activation patterns. The layer-wise observations are consistent with progressively more abstract feature representations as depth increases.

## H. ROC and Precision-Recall Analysis

Class-wise Receiver Operating Characteristic (ROC) and Precision-Recall (PR) curves were evaluated using a one-vs-rest decomposition on the 10,933-image image-level held-out test set. Figure 23 presents the ROC and PR curves for the four diagnostic classes.

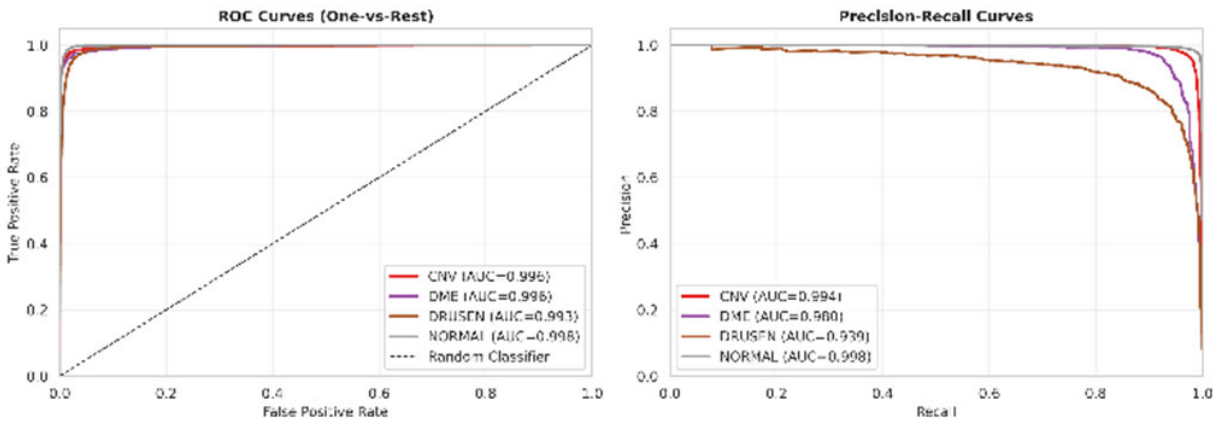

Fig. 23. One-vs-rest ROC and Precision-Recall curves for the four retinal OCT classes on the image-level held-out test set.

**ROC analysis**: All four classes remained close to the upper-left region of ROC space. The resulting one-vs-rest ROC-AUC values were 0.996 for CNV, 0.996 for DME, 0.993 for Drusen, and 0.998 for Normal. The Drusen class therefore produced the lowest ROC-AUC among the four classes, although the difference was small in absolute terms. Its specificity at the reported multiclass operating point was 0.9716.

**PR analysis**: The PR curves revealed greater class-conditional differences. CNV and Normal maintained precision above approximately 0.95 across most of the recall range, while DME exhibited a more moderate decline as recall increased. Drusen showed the strongest precision degradation as recall approached higher operating regions. The Average Precision values were approximately 0.99 for CNV, 0.98 for DME, 0.94 for Drusen, and 0.999 for Normal.

The lower Average Precision observed for Drusen is consistent with the class-level precision and confusion-matrix results reported above. In particular, the substantial number of false-positive Drusen assignments contributes directly to the reduced precision-recall performance of that class. Thus, ROC-AUC alone does not fully capture the class-conditional asymmetry observed for Drusen in the present experiment.

Across the evaluated metrics, the principal class-conditional limitation was consistently associated with Drusen. Drusen achieved high sensitivity but lower precision and F1 than the other classes, and the same asymmetry was reflected in the confusion matrix and the lower Average Precision observed in the one-vs-rest Precision-Recall analysis. These results identify Drusen-related false-positive assignments as the principal residual error pattern for ED126-A.

## V. Discussion

The multi-faceted evaluation provides an empirical assessment of both research questions: whether a purpose-built CNN trained from scratch provides a useful benchmark baseline for four-class OCT classification (RQ1), and which pathological classes constitute the principal sources of residual classification error (RQ2).

ED126-A provides a documented from-scratch reference condition for four-class retinal OCT classification. On the image-level held-out test set, the model achieved 95.03% accuracy, macro-F1 of 0.9205, and weighted-F1 of 0.9521. The aggregate results, however, mask substantial class-conditional asymmetry. Drusen achieved the highest recall among the four classes (0.9605) but the lowest precision (0.7493) and F1-score (0.8419). The confusion matrix indicates that this asymmetry is driven primarily by false-positive Drusen assignments, particularly CNV → Drusen (176 cases) and Normal → Drusen (96 cases). The lower Drusen Average Precision observed in the one-vs-rest PR analysis is consistent with the broader difficulty of maintaining precision across the recall range, despite the high ROC-AUC of 0.993. Collectively, these results indicate that aggregate accuracy and ROC-AUC alone do not fully characterize the residual error structure of this baseline. For the compact-CNN studies considered in this review, the reported results emphasize aggregate performance, limiting direct assessment of class-conditional failure modes from the published metrics [3], [5], [13].

The principal empirical finding is a pronounced precision-recall asymmetry for the Drusen class (F1 = 0.8419, precision = 0.7493, recall = 0.9605). The observed error pattern, comprising Normal → Drusen (96, 53.3% of Normal-class errors), CNV → Drusen (176, 65.7% of CNV errors, 61.8% of Drusen false positives), and DME → Normal (33), is consistent with ambiguity in the learned representation and decision boundaries, although the present experiment does not isolate representational effects from other architectural and optimization factors. Four complementary analyses provide qualitatively consistent observations regarding the residual error structure: (i) t-SNE shows a predominantly compact Drusen cluster with localized overlap toward neighboring class regions, with misclassifications forming contiguous boundary-region blocks rather than scattered outliers; (ii) the winner-runner-up margin distribution indicates that a non-trivial subset of predictions retain substantial competing-class probability; (iii) Grad-CAM on misclassified CNV shows less focal, multi-lobed saliency; (iv) under the balanced-class-weight configuration, Drusen recall was 0.9605, the highest among the four classes, while the observed Drusen-related false-positive assignments remained substantial. One possible architectural factor is the progressive spatial reduction from $224 \times 224$ at the input to $14 \times 14$ after Block 4. This reduction could limit the preservation of fine-grained spatial information relevant to small morphological structures; however, this hypothesis is not tested by the present experiment and cannot be separated from other architectural and optimization factors. In the test set, 285 of 1,137 predictions assigned to the Drusen class were false positives, corresponding to a false-positive proportion of 25.1% among Drusen predictions.

The per-class ROC-AUC values cluster tightly in [0.993, 0.998] (macro ROC-AUC ≈ 0.996), which alone would suggest uniformly excellent discrimination, yet Average Precision exposes the concealed asymmetry: Drusen exhibited an AP of approximately 0.94 despite a ROC-AUC of 0.993, illustrating that the two threshold-independent summaries provide different views of class discrimination under the one-vs-rest formulation. The difference between the ROC and PR summaries is consistent with the class-conditional effects of the one-vs-rest negative-class composition: the one-vs-rest ROC-AUC is evaluated against a heterogeneous negative-class pool that includes the substantially larger Normal class. Consequently, ROC-AUC may provide a less sensitive view of the Drusen false-positive burden than the corresponding PR analysis. The PR analysis is particularly sensitive to the prevalence of the positive class and the distribution of false-positive predictions and reflects the observed 96 Normal → Drusen and 176 CNV → Drusen false-positive assignments in the corresponding precision-recall behavior. This pattern is consistent with the metric-level manifestation of the same residual error structure reflected in the F1 results and t-SNE visualization. OctNET [3], the lightweight CNN evaluated for adversarial robustness [5], and OCT-Caps [17] report aggregate accuracy without per-class PR analysis. In contrast, the present analysis shows that ROC-AUC values of at least 0.993 can coexist with substantially lower operating-point precision, reaching 0.7493 for Drusen. Per-class PR analysis with AP reporting is therefore recommended when evaluating class-conditional performance on this benchmark, particularly for morphologically ambiguous classes such as Drusen.

### A. Limitations and Future Work

Five limitations constrain the interpretation of the present findings. (1) Single-dataset evaluation: the Kermany cohort exhibits substantial class imbalance and geographically concentrated sampling; cross-dataset validation on an independent OCT cohort is therefore required. (2) Restricted Grad-CAM sampling: the presented qualitative analysis is limited to CNV examples; class-stratified analyses of the principal Normal → Drusen and CNV → Drusen error channels are needed. (3) Single-seed training: all reported metrics are point estimates from one training run; replication across multiple independent random seeds would provide a more robust estimate of variability and enable confidence-interval reporting. (4) Unresolved architectural effects: whether increased depth or alternative architectural components can reduce the observed Drusen-related confusion remains an open empirical question, and the present experiment cannot isolate the marginal contribution of ImageNet-pretrained initialization. (5) No explicit calibration analysis was performed; consequently, the calibration properties of the softmax outputs remain unresolved. Post-hoc calibration, such as temperature scaling, is a relevant direction for future evaluation.

Because the experimental partition was performed at the image rather than patient level, related images from the same patient may be represented across training, validation, and test subsets. Consequently, the reported test performance may overestimate performance under patient-independent evaluation, and the results should not be interpreted as evidence of clinical generalization.

The forward research agenda is based on this baseline and includes: transfer learning under identical protocol, attention/multi-scale enrichment (CBAM, FPN) targeting the observed Drusen-related overlap region in the t-SNE projection, and threshold-optimized deployment calibration.

## VI. Conclusion

This study presents ED126-A as a compact, purpose-built CNN baseline for four-class retinal OCT classification, trained from random initialization under a documented and fixed experimental protocol. On the image-level held-out test set, ED126-A achieved 95.03% accuracy, a macro-F1 of 0.9205, and a weighted-F1 of 0.9521.

The principal empirical finding was substantial class-conditional asymmetry, particularly for Drusen. Although Drusen recall reached 0.9605, precision was 0.7493, and F1 was 0.8419. The confusion matrix identified CNV → Drusen and Normal → Drusen as the dominant false-positive channels, demonstrating that aggregate accuracy alone does not adequately characterize the residual error structure of the baseline.

The complementary confidence and representation analyses were consistent with localized ambiguity in the corresponding decision regions, but they do not establish a causal architectural explanation for the observed errors. In particular, the contribution of depth, spatial resolution, pretrained initialization, and other architectural factors remains unresolved by the present experiment.

The results support the use of ED126-A as a reference baseline for subsequent controlled experiments and motivate controlled comparisons of transfer learning, increased depth, attention, multi-scale feature fusion, and calibration under an otherwise matched experimental protocol.

## Declarations

**Competing Interests**: The authors declare no competing interests.

**Funding Information**: No funding was received for this study.

**Author Contribution**:

**Shkëlqim Sherifi***: Conceptualization, methodology, software, data curation, formal analysis, investigation, writing - original draft, writing - review & editing, visualization, and project administration.

**Festim Halili and Merita Kasa-Halili**: Methodology, formal analysis and writing - review.

**Data Availability Statement**: The data supporting this study are publicly available at:
https://github.com/shkelqimsherifi/CNN-Based-Retinal-OCT-Classification-From-Scratch-Baseline-ED126-A.git

The retinal OCT dataset used in this study is publicly available through Kaggle: https://www.kaggle.com/datasets/anirudhcv/labeled-optical-coherence-tomography-oct and the original Kermany et al. dataset deposited in Mendeley Data: https://doi.org/10.17632/rscbjbr9sj.3.

**Research Involving Human and Animals:** This study involved secondary analysis of publicly available retinal OCT images. No new participants were recruited or contacted, and no animal experiments were conducted.

**Informed Consent**: No new informed consent was obtained for this secondary analysis of previously collected, publicly available images. Informed-consent procedures for the original data collection are governed by the original dataset documentation.

## References

[1] K. Du *et al.*, "Detection of Disease Features on Retinal OCT Scans Using RETFound," *Bioengineering*, vol. 11, no. 12, p. 1186, Dec. 2024, doi: 10.3390/bioengineering11121186.

[2] P. K. Karn and W. H. Abdulla, "Precision Segmentation of Subretinal Fluids in OCT Using Multiscale Attention-Based U-Net Architecture," *Bioengineering*, vol. 11, no. 10, p. 1032, Oct. 2024, doi: 10.3390/bioengineering11101032.

[3] A. P. Sunija, S. Kar, S. Gayathri, V. P. Gopi, and P. Palanisamy, "OctNET: A Lightweight CNN for Retinal Disease Classification from Optical Coherence Tomography Images," *Comput. Methods Programs Biomed.*, vol. 200, p. 105877, Mar. 2021, doi: 10.1016/j.cmpb.2020.105877.

[4] S. Pang *et al.*, "A novel approach for automatic classification of macular degeneration OCT images," *Sci. Rep.*, vol. 14, no. 1, p. 19285, Aug. 2024, doi: 10.1038/s41598-024-70175-2.

[5] M. Bhandari, T. B. Shahi, and A. Neupane, "Evaluating Retinal Disease Diagnosis with an Interpretable Lightweight CNN Model Resistant to Adversarial Attacks," *J. Imaging*, vol. 9, no. 10, p. 219, Oct. 2023, doi: 10.3390/jimaging9100219.

[6] X. Huang *et al.*, "GABNet: global attention block for retinal OCT disease classification," *Front. Neurosci.*, vol. 17, Jun. 2023, doi: 10.3389/fnins.2023.1143422.

[7] X. Chen *et al.*, "An artificial intelligence cloud platform for OCT-based retinal anomalies screening system in real clinical environments," *Npj Digit. Med.*, vol. 8, no. 1, p. 559, Aug. 2025, doi: 10.1038/s41746-025-01959-7.

[8] M. Badar, M. Haris, and A. Fatima, "Application of deep learning for retinal image analysis: A review," *Comput. Sci. Rev.*, vol. 35, p. 100203, Feb. 2020, doi: 10.1016/j.cosrev.2019.100203.

[9] A. Nawaz, T. Ali, G. Mustafa, M. Babar, and B. Qureshi, "Multi-Class Retinal Diseases Detection Using Deep CNN With Minimal Memory Consumption," *IEEE Access*, vol. 11, pp. 56170–56180, 2023, doi: 10.1109/ACCESS.2023.3281859.

[10] T. A. Soomro *et al.*, "Deep Learning Models for Retinal Blood Vessels Segmentation: A Review," *IEEE Access*, vol. 7, pp. 71696–71717, 2019, doi: 10.1109/ACCESS.2019.2920616.

[11] İ. Kayadibi, G. E. Güraksın, and U. Köse, "A Hybrid R-FTCNN based on principal component analysis for retinal disease detection from OCT images," *Expert Syst. Appl.*, vol. 230, p. 120617, Nov. 2023, doi: 10.1016/j.eswa.2023.120617.

[12] S. Sherifi, "Deep Learning Architectures for Visual Representation and Image Classification: From Multilayer Perceptrons to CNNs and Hybrid Models - A Technical Survey," Sep. 01, 2026, *Social Science*

*Research Network, Rochester, NY*: 7467760. doi: 10.2139/ssrn.7467760.

[13] R. R. Teegavarapu *et al.*, “Evaluation of Convolutional Neural Networks (CNNs) in Identifying Retinal Conditions Through Classification of Optical Coherence Tomography (OCT) Images,” *Cureus*, vol. 17, Jan. 2025, doi: 10.7759/cureus.77109.

[14] Z. Ma, Q. Xie, P. Xie, F. Fan, X. Gao, and J. Zhu, “HCTNet: A Hybrid ConvNet-Transformer Network for Retinal Optical Coherence Tomography Image Classification,” *Biosensors*, vol. 12, no. 7, p. 542, Jul. 2022, doi: 10.3390/bios12070542.

[15] S. Sherifi, “CNN-Based Retinal OCT Classification: A Scoping Review and Research Gap Analysis,” Jun. 01, 2026, *Social Science Research Network, Rochester, NY*: 7606098. doi: 10.2139/ssrn.7606098.

[16] T. S. Apon, M. M. Hasan, A. Islam, and M. G. R. Alam, “Demystifying Deep Learning Models for Retinal OCT Disease Classification using Explainable AI,” in *2021 IEEE Asia-Pacific Conference on Computer Science and Data Engineering (CSDE)*, Dec. 2021, pp. 1–6. doi: 10.1109/CSDE53843.2021.9718400.

[17] T. Tsuji *et al.*, “Classification of optical coherence tomography images using a capsule network,” *BMC Ophthalmol.*, vol. 20, p. 114, Mar. 2020, doi: 10.1186/s12886-020-01382-4.

[18] “EyeNet.” Accessed: Mar. 09, 2026. [Online]. Available: https://kaggle.com/code/nikolasgegenava/eyenet

[19] “Awesome-Medical-Dataset/resources/STARE.md at main · openmedlab/Awesome-Medical-Dataset.” Accessed: Mar. 09, 2026. [Online]. Available: https://github.com/openmedlab/Awesome-Medical-Dataset/blob/main/resources/STARE.md

[20] “DRIVE: Digital Retinal Images for Vessel Extraction Dataset — bob.db.drive 1.0.1 documentation.” Accessed: Mar. 09, 2026. [Online]. Available: https://www.idiap.ch/software/bob/docs/bob/bob.db.drive/stable/index.html

[21] “Labeled Optical Coherence Tomography (OCT).” Accessed: Feb. 22, 2026. [Online]. Available: https://www.kaggle.com/datasets/anirudhcv/labeled-optical-coherence-tomography-oct

[22] D. Kermany, K. Zhang, and M. Goldbaum, “Large Dataset of Labeled Optical Coherence Tomography (OCT) and Chest X-Ray Images,” Jun. 2018, doi: 10.17632/rscbjbr9sj.3.

[23] R. R. Selvaraju, M. Cogswell, A. Das, R. Vedantam, D. Parikh, and D. Batra, “Grad-CAM: Visual Explanations from Deep Networks via Gradient-based Localization,” *Int. J. Comput. Vis.*, vol. 128, no. 2, pp. 336–359, Feb. 2020, doi: 10.1007/s11263-019-01228-7.

## VII. Appendix A

### A.1 Prototype Overview

A Windows desktop prototype was developed to demonstrate local inference with the trained ED126-A model. The prototype was implemented using the Qt framework for Python and provides a graphical interface for loading a trained model, selecting OCT images from local storage, executing inference, and visualizing the resulting class probabilities. The application is intended as a research and demonstration interface and is not a clinically validated diagnostic system.

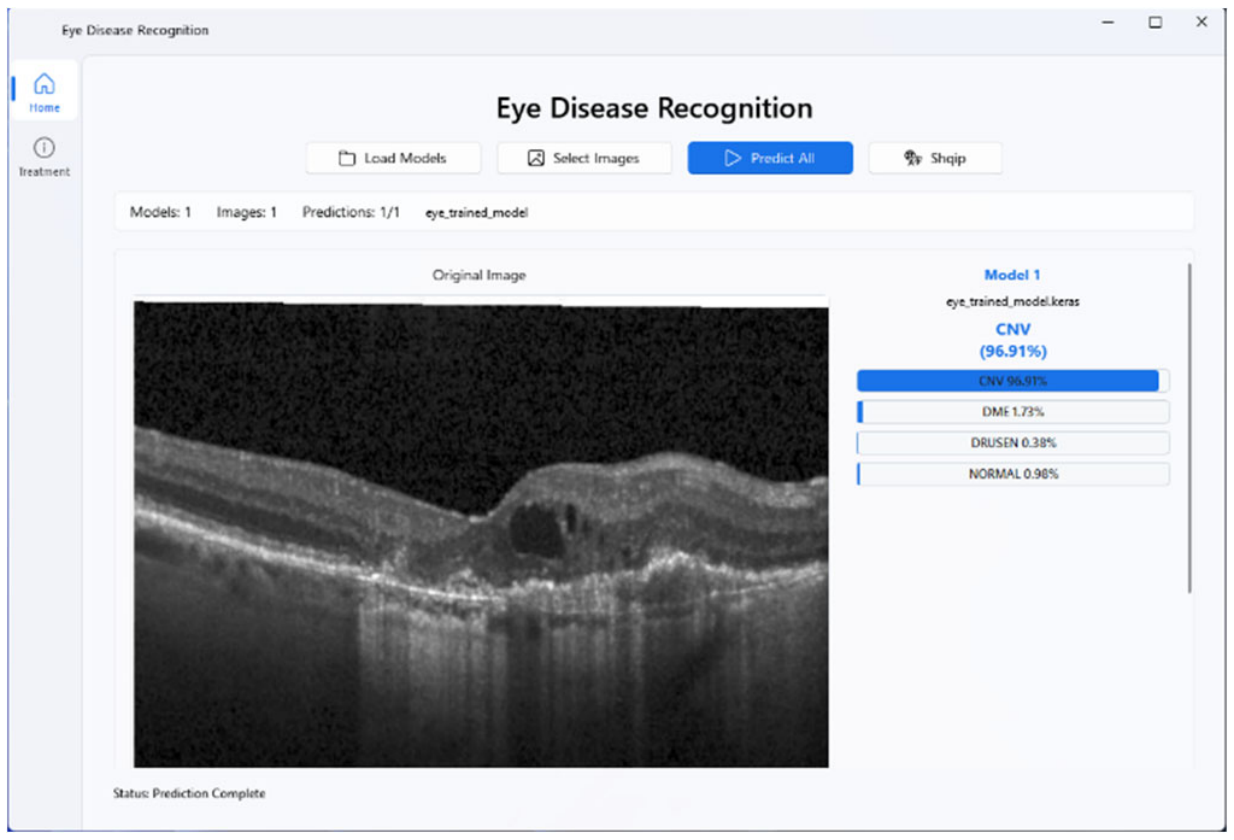

Fig. A1. ED126-A desktop application interface for local OCT image inference and probability visualization.

The prototype uses the same four target categories as the experimental model: Choroidal Neovascularization (CNV), Diabetic Macular Edema (DME), Drusen, and Normal. The interface displays the predicted class together with the model-generated probability distribution across the four categories. The example shown in Fig. A1 illustrates an input OCT image for which CNV received the highest predicted probability.

### A.2 Inference Workflow

The inference workflow consists of sequential model-loading, input-validation, prediction, and result-presentation stages, as illustrated in Fig. A2. The user first selects a trained model file. The application performs a model-loading and validity check before inference is enabled. The user then selects one or more OCT images from local storage, after which the input files are checked for readability and compatibility with the configured preprocessing pipeline.

Following successful validation, the application performs a forward pass through the loaded model and displays the predicted class probabilities. The interface also provides controls for loading a model, selecting one or more images, executing inference, and switching the display language. Status information is presented within the application interface to indicate the current model and inference state.

### A.3 Prediction Visualization

The prediction-results panel displays the input OCT image together with the predicted class and the associated class-probability distribution. This presentation allows the user to inspect the model output across all four classes rather than viewing only the top-ranked prediction.

The probability visualization is intended to expose the model output used by the classification decision. These values should be interpreted as model-generated softmax outputs and should not be regarded as clinically calibrated probabilities.

### A.4 Prototype Scope and Disclaimer

The desktop application is a software prototype developed for research and demonstration purposes. Its interface, local inference capability, and visualization functions do not constitute clinical validation, prospective clinical evaluation, or regulatory approval.

The Treatment tab, where present, is an interface prototype for displaying externally defined informational content. It is not intended to provide validated clinical treatment recommendations or to substitute for professional medical judgment. The present study evaluates the ED126-A classification model on OCT images. It does not evaluate the safety, effectiveness, or clinical utility of treatment recommendations generated or displayed by the software prototype.

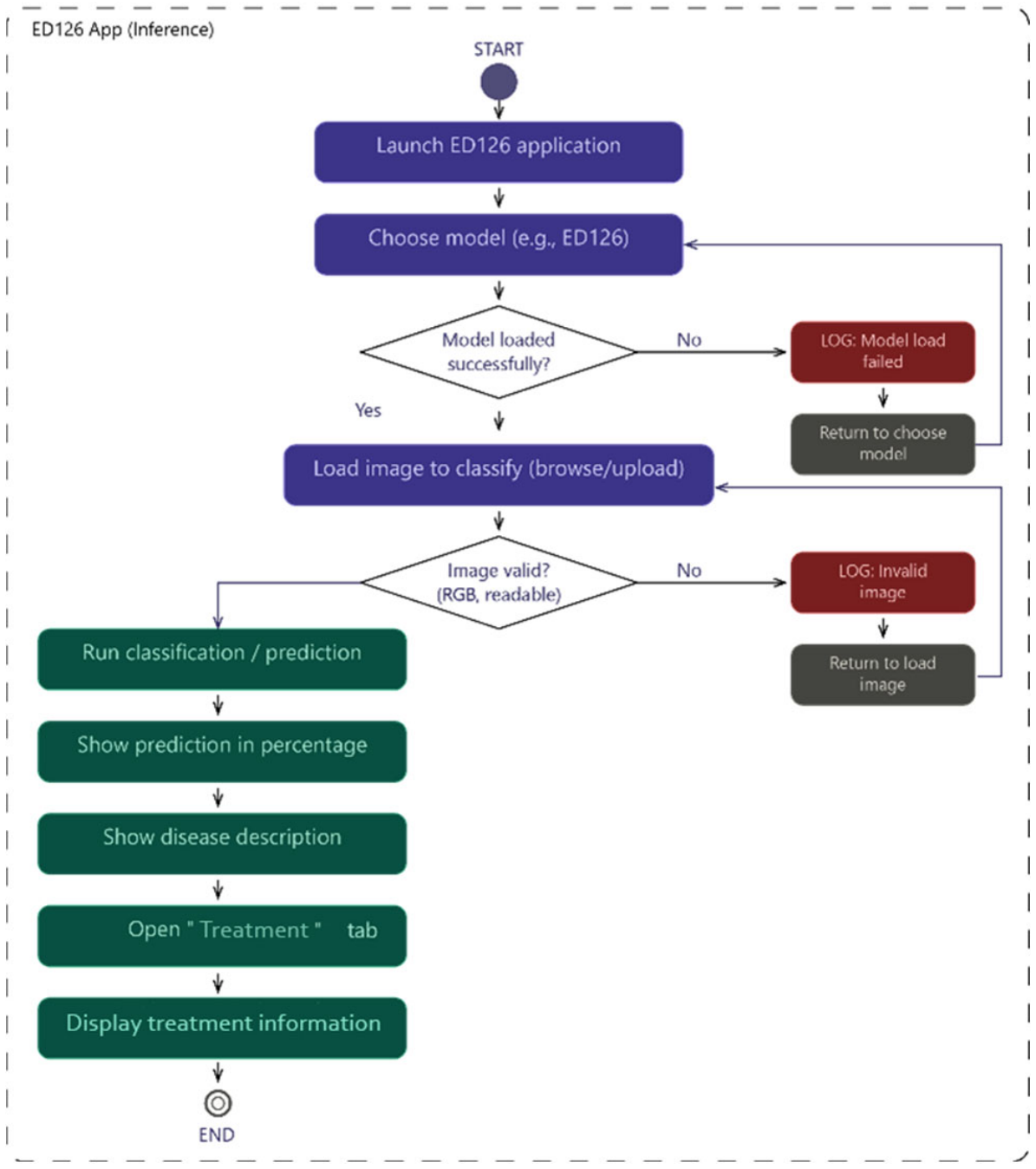

Fig. A2. Inference workflow of the ED126-A desktop application.

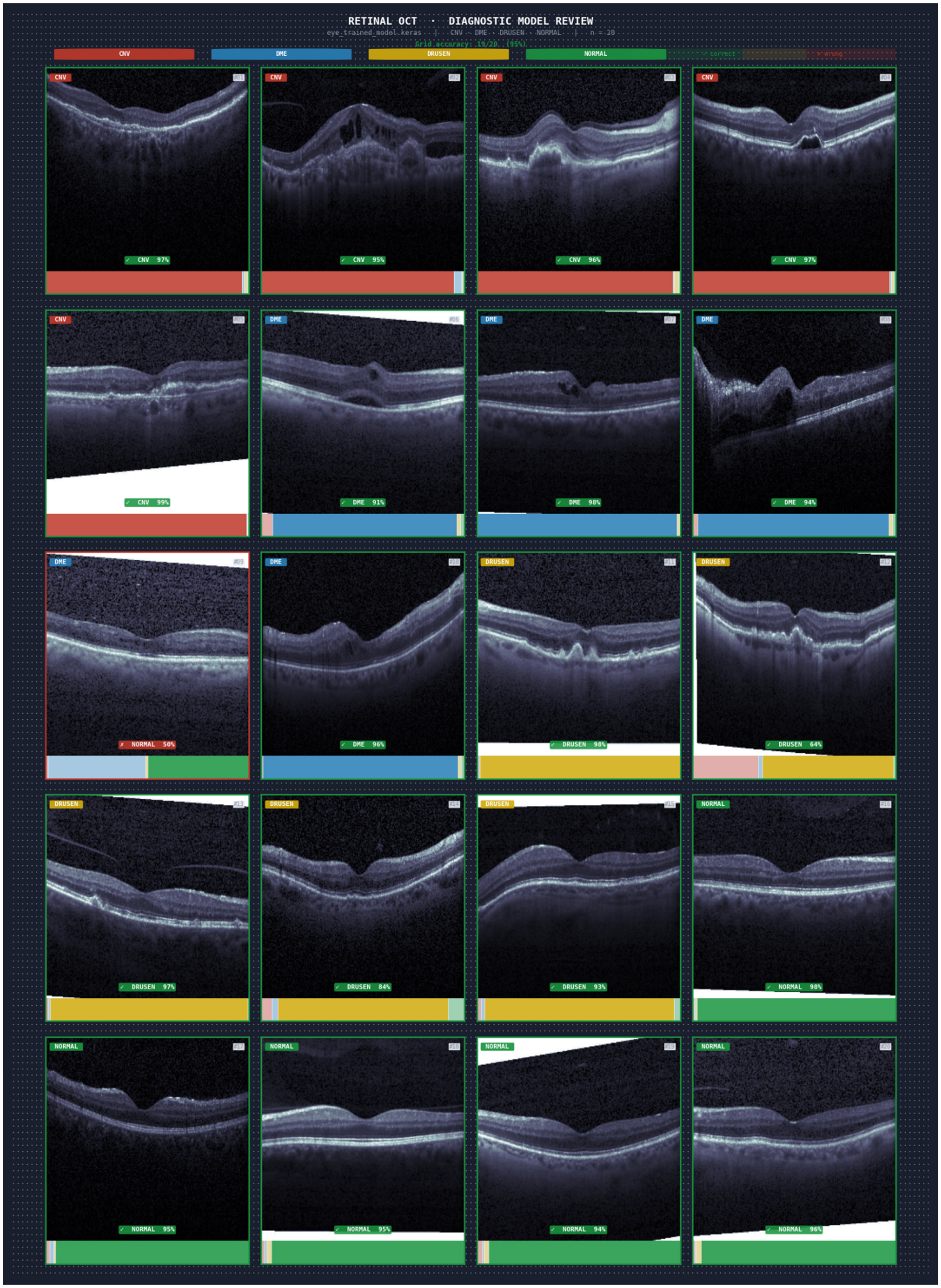

Fig. A3. ED126-A model review interface showing predicted class probabilities for multiple example OCT images.